\def\year{2022}\relax
\documentclass[letterpaper]{article} 
\usepackage{aaai22}  
\usepackage{times}  
\usepackage{helvet}  
\usepackage{courier}  
\usepackage[hyphens]{url}  
\usepackage{graphicx} 
\usepackage{natbib}  
\usepackage{caption} 
\DeclareCaptionStyle{ruled}{labelfont=normalfont,labelsep=colon,strut=off} 
\usepackage{algorithm}
\usepackage{algorithmic}
\usepackage{subcaption}
\usepackage{multirow}
\newtheorem{definition}{Definition} 
\usepackage{amsmath} 
\usepackage{amssymb}

\usepackage{newfloat}
\usepackage{listings}
\floatstyle{ruled}
\newfloat{listing}{tb}{lst}{}
\floatname{listing}{Listing}

\title{An Unsupervised Way to Understand Artifact Generating Internal Units \\in Generative Neural Networks}
\author {
    Haedong Jeong\textsuperscript{\rm 1,2},
    Jiyeon Han\textsuperscript{\rm 2},
    Jaesik Choi\textsuperscript{\rm 2,3}\thanks{Corresponding Author}
}
\affiliations {
    \textsuperscript{\rm 1} Ulsan National Institute of Science and Technology (UNIST), South Korea\\
    \textsuperscript{\rm 2} Korea Advanced Institute of Science and Technology (KAIST), South Korea\\
    \textsuperscript{\rm 3} INEEJI, South Korea\\
    \{haedong.jeong, j.han, jaesik.choi\}@kaist.ac.kr
}

\begin{document}

\maketitle

\begin{abstract}
Despite significant improvements on the image generation performance of Generative Adversarial Networks (GANs), generations with low visual fidelity still have been observed. As widely used metrics for GANs focus more on the overall performance of the model, evaluation on the quality of individual generations or detection of defective generations is challenging. While recent studies try to detect featuremap units that cause artifacts and evaluate individual samples, these approaches require additional resources such as external networks or a number of training data to approximate the real data manifold. 
In this work, we propose the concept of \textit{local activation}, and devise a metric on the local activation to detect artifact generations without additional supervision.
We empirically verify that our approach can detect and correct artifact generations from GANs with various datasets. Finally, we discuss a geometrical analysis to partially reveal the relation between the proposed concept and low visual fidelity.
\end{abstract}

\section{Introduction}
Since the adversarial generative training scheme \cite{goodfellow2014generative} emerged, deep generative neural networks (DGNNs) have shown incredible performance on image generation tasks. From recent research with generative adversarial networks (GANs), various structures and training strategies \cite{brock2018large,karras2018progressive,DBLP:conf/iclr/MiyatoKKY18,karras2019style} have been proposed to overcome the weaknesses of the adversarial training scheme (e.g., unstable training) and to accelerate the improvements of visual fidelity of the generations.

Despite significant improvements, models sometimes present undesirable outcomes such as perceptually defective generations called \textit{artifacts}. Various metrics have been suggested to evaluate the performance of a generator \cite{salimans2016improved,conf/nips/HeuselRUNH17,conf/nips/SajjadiBLBG18,NEURIPS2019_0234c510}. However, it is non-trivial to evaluate the visual fidelity of each individual sample because existing metrics mainly focus on the distributional difference between the real dataset and the generations in the feature manifold. A research, which uses the nearest neighbor based similarity in the feature manifold \cite{NEURIPS2019_0234c510}, was proposed as an alternative to estimate the quality of individual generations. Although this method has shown effectiveness in evaluation, it requires a huge amount of real data and an external network for feature embedding to make the scoring process reliable.

A few studies have been conducted to understand the internal generation mechanism of GANs to detect or correct the individual generations with low visual fidelity. In GAN Dissection \cite{bau2018visualizing}, the authors identify the defective units that mainly cause artifacts based on a set of generations on which a featuremap unit is highly activated. The authors further improve the fidelity of individual generations by zero-ablating the detected featuremap units. A similar approach trains an external classifier to extract the region with low visual fidelity in the individual generations and identifies internal units related to the extracted region \cite{tousi2021automatic}. On the other hand, manipulation of the latent code based on the binary linear classifier has been proposed to correct the artifact \cite{interfacegan}. While these approaches can be utilized to evaluate the fidelity of individual samples, they still require additional resources such as a human annotation process.

\begin{figure*}[t!]
    \centering
    \includegraphics[width=0.95\textwidth]{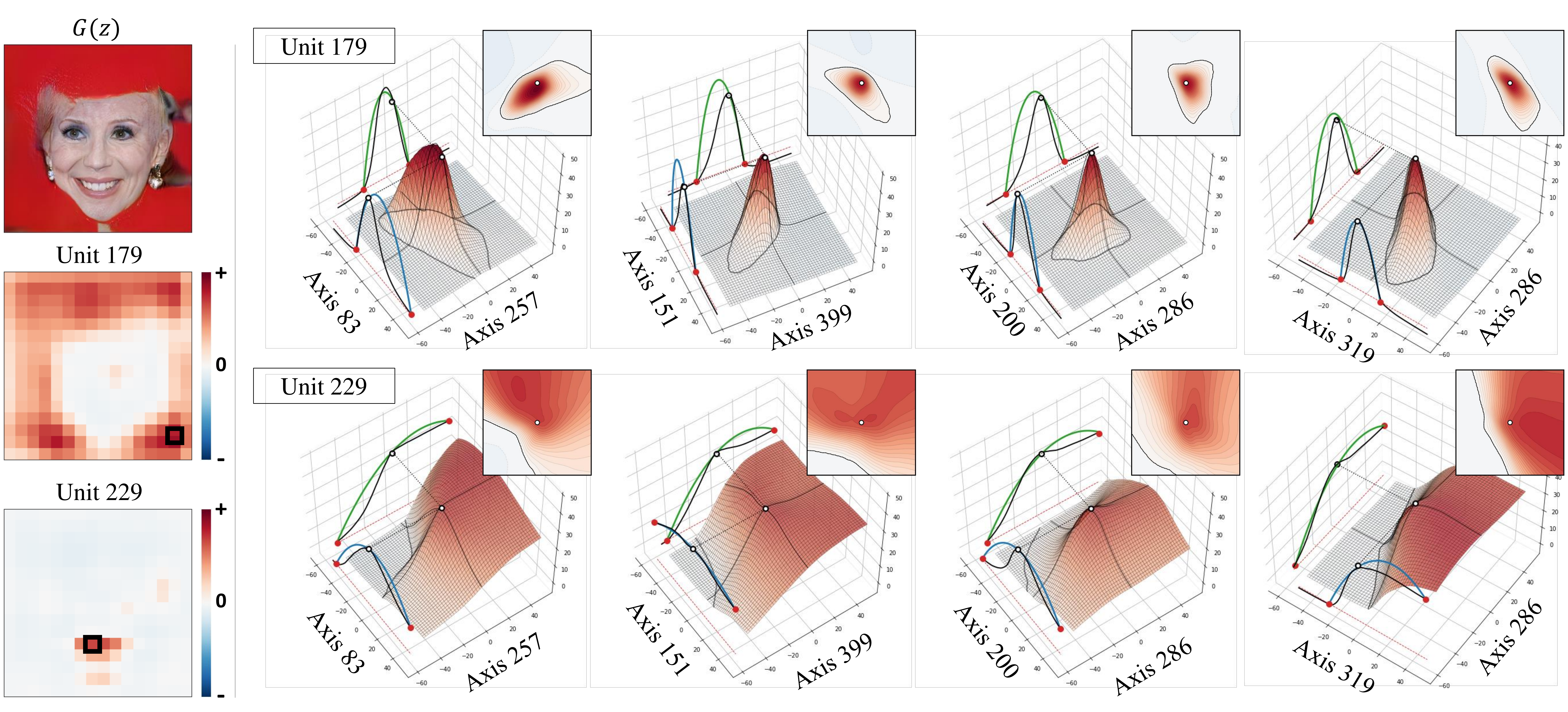}
    \caption{An illustrative example of the locally activated neuron in PGGAN model trained on CelebA-HQ. We manually select two featuremap units in layer 6 ($\in \mathbb{R}^{512\times16\times16}$) which are expected to relate to low visual fidelity (red background in $G(z)$) and mouth. (Left) The black box in each featuremap unit means the spatial information of the selected neuron. (Right) The sections of activation patterns in the latent space. The black solid line in each side means the activation pattern in the corresponding axis. The axis is randomly selected for the visualization. The red dots indicate the change points (Definition \ref{def:cp}) and green/blue lines are the approximated curvature of local activation for each axis. The neuron in unit 179 has more locally activated pattern compared to the activation pattern of the neuron in unit 229.}
    \label{fig:la_eg}
\end{figure*}

In this paper, we propose the concept of \textit{local activation} to detect and correct the artifact generations in an unsupervised manner. We also discuss a geometrical analysis to partially investigate the relation between the local activation and low visual fidelity of individual generations.
The main advantages of our method are twofold: (1) external networks or supervisions are unnecessary to detect and correct artifact generations. The evaluation is performed solely on the target generator by using the internal property for scoring, and (2) the proposed approach can be applicable to various structures of GANs for evaluating the visual fidelity, because the proposed approach is based on neurons that are the common basic components of neural networks. We experimentally verify that our method can detect and correct artifact effectively on PGGAN \cite{karras2018progressive}, and StyleGAN2 \cite{karras2020analyzing} with various datasets.

\section{Related Work}
\textbf{Deep Generative Neural Networks} DGNNs are the models which approximate the input distribution given a target with neural networks. Representative architectures include variational autoencoder (VAE) \cite{kingma2013auto}, neural language models \cite{peters-etal-2018-deep,kenton2019bert,brown2020language} and GANs. In particular, the adversarial training between a generator and a discriminator \cite{goodfellow2014generative} has shown impressive performance in image generation \cite{karras2018progressive,karras2019style,karras2020analyzing,brock2018large}.



\textbf{Analysis for Interior Mechanism of GANs} GAN Dissection \cite{bau2018visualizing} proposes a framework to investigate the generative role of each featuremap unit in GANs. It is shown that artifact generations can be improved by ablating units that are related to artifact generations. Another work \cite{interfacegan} trains a linear classifier based on artifact-labeled data and removes artifacts by moving the latent code over the trained hyperplane. A sampling method with the trained generative boundaries was suggested to explain shared semantic information in the generator \cite{jeon2020efficient}. Classifier-based defective internal featuremap unit identification was devised \cite{tousi2021automatic}. The authors increase the visual fidelity by sequentially controlling the generation flow of the identified units. 
Analyses for latent space of the generator were also performed to manipulate the semantic of the generation \cite{peebles2020hessian,harkonen2020ganspace}. Our work focuses more on the generation process and the relation between defective generation and the internal characteristics connected from the latent space.

\textbf{Metric for Generative Model} 
Various metrics have been proposed to evaluate the performance of generative models and each properties are well-summarized in \cite{borji2019pros}. Although Fréchet Inception Distance (FID) \cite{conf/nips/HeuselRUNH17} and Inception Score (IS) \cite{salimans2016improved} have shown robustness to image distortion, they sometimes assign high scores for generations with low visual fidelity. Precision and Recall (P\&R) is a surrogate metric to quantify mode dropping and inventing based on training data \cite{conf/nips/SajjadiBLBG18,NEURIPS2019_0234c510}. The authors also devised Realism Score (RS) to evaluate the visual fidelity of individual samples by comparing feature embeddings with training data. Perceptual path length (PPL) is another metric that quantifies the smoothness of the latent space with a hypothesis that the region in the latent space for defective generations has a small volume \cite{karras2019style}.


\section{Locally Activated Neurons in GANs}
In this section, we present our main contribution, the concept of \textit{local activation} and its relation with low visual fidelity for individual generations. 
From previous research \cite{bau2018visualizing,jeon2020efficient,tousi2021automatic}, we can presume that each internal featuremap unit in the generator handles a specific object (e.g., tree, glasses) for the final generation. In particular, an artifact that has low visual fidelity can also be considered as a type of object.
Thus, it is possible to identify the units causing low visual fidelity. To expand these observations, we focus on neurons as the basic component of a featuremap unit.

\begin{figure}[t!]
    \centering
    \includegraphics[width=0.97\columnwidth]{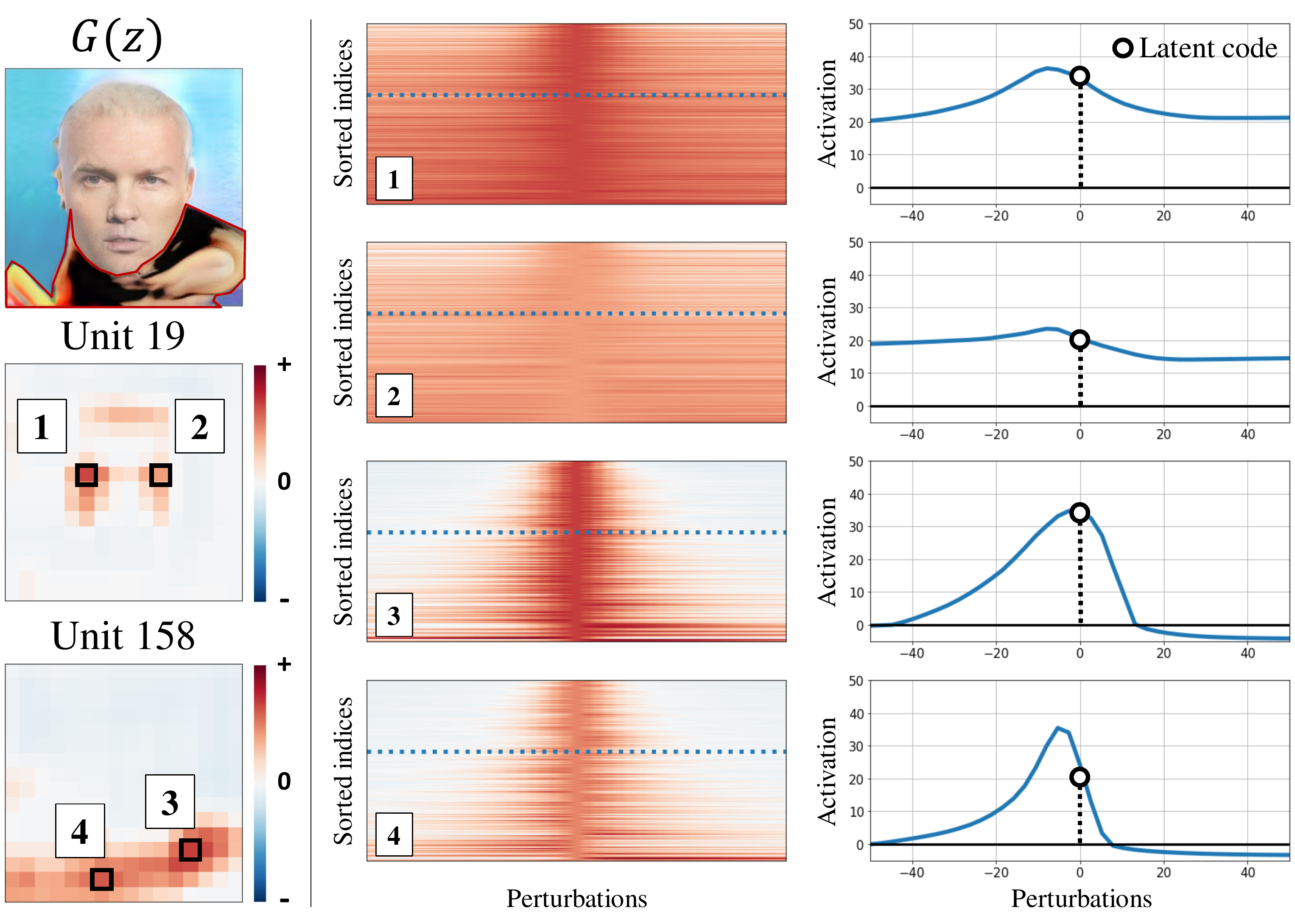}
    \caption{Activation patterns of the neurons in the manually selected featuremap units for the given latent code $z$. The middle column represents the heatmaps of the activation patterns around $z$ where each row corresponds to each axis of the latent space. The rows are sorted by the local activation in the descending order from top to bottom. The right-most column represents the activation patterns for a specific axis (the dotted rows in the middle column) in the 2D representation. The neurons related to the defective region are more locally activated.}
    \label{fig:la_2d}
\end{figure}

\subsection{Quantification of Local Activation}\label{sec:quant}
We observe the neurons that correspond to the artifact region often show a bounded activation pattern. Figure \ref{fig:la_eg} shows one example of artifact generation and two featuremap units in PGGAN. Unit 179 is highly correlated to the artifact region (red background) in the generation and unit 229 corresponds to the mouth part, which shows high visual fidelity. The right side of Figure \ref{fig:la_eg} shows 3D representations of the activation patterns in the latent space for two neurons from unit 179 and unit 229, respectively. The neuron from unit 229 shows high activation over a large area in the latent space. In contrast, the neuron from unit 179, which corresponds to the artifact region, shows high activation only in a restricted area across various pairs of axes.
Figure \ref{fig:la_2d} further supports how the activation patterns are different between the artifact-related neurons and the normal neurons. In the second column, activation patterns of the artifact-related neurons (neurons 3 and 4) are more sharply concave across the latent axes compared to the normal neurons (neurons 1 and 2). The concave shape of the activation pattern suggests that the activation is bounded and concentrated around the given latent code. 

From the observations in Figures \ref{fig:la_eg} and \ref{fig:la_2d}, we suspect that the bounded activation pattern may be related to the visual fidelity of a generation. We call this bounded activation pattern \textit{local activation} and the corresponding neuron a \textit{locally activated neuron}.
However, it is non-trivial to exactly quantify the local activation for an internal neuron in the latent space, because (1) commonly used generators have high dimensional latent space, and (2) the activation pattern forms a highly non-convex shape in the latent space. To mitigate these problems, we approximate the curvature of the local activation pattern with a line search for each latent dimension within the empirical search bound. 

Let the generator $G$ with $L$ layers be $G(z)=g_{L}(g_{L-1}(\cdots (g_{1}(z)))) = g_{L:1}(z)$, where $z$ is a vector in the latent space $\mathcal{Z}\subset \mathbb{R}^{D_z}$, $g_l(h_{l-1})=\sigma(w_{g_l}^\top h_{l-1})$, $h_{l-1}=g_{l-1:1}(z)$, and $\sigma(\cdot)$ is an activation function such as LeakyReLU or ReLU\footnote{There are various activation functions for deep neural networks, we only consider LeakyReLU and ReLU function in this paper.}. One can express the bias of each layer in the homogeneous representation with this unified equation by applying an additional dimension for the bias. 
For the $i$-th neuron $g_{l:1}^i(z)$ of $g_{l:1}(z)$, we can obtain an activation pattern with the line search over the perturbation range for each latent dimension and we define the left/right change points for each activation pattern as follows. 

\begin{definition}[\textbf{Change Point}]
\label{def:cp}
Let the given latent code be $z_0$, the dimension index be $d\in\{1,...,D_z\}$, the search bound be $R>0$, and the canonical basis be $\textbf{e}_d=(0,...,0,1,0,...,0)^\top$ with the nonzero component at position $d$. For the set of change point $P=\{r|g_{l:1}^i(z_0+r\cdot\textbf{e}_d)=0\}\cup \{-R,R\}$ for $r\in[-R,R]$, the right and the left change points of $i$-th neuron at the $l$-th layer are defined respectively as,
\begin{equation}
    {}^rp =  \min_{\forall r\in P; \\ r\ge0}(r) \quad\text{and}\quad {}^lp = \max_{\forall r\in P; \\ r\le0}(r).  
\end{equation}

\end{definition}

We note that if there are no points where activation signs are changed, the search bounds are considered as the change points by Definition 1. We approximate the curvature of the local activation by computing the curvature (the coefficient of the second degree term) of the quadratic approximation of three points (the left/right change points and the given latent code $z_0$) for each latent axis, and averaging over the latent dimensions. The green and blue curves in Figure \ref{fig:la_eg} illustrate the approximated quadratic functions for quantifying the local activation. 

\begin{definition}[\textbf{Curvature of Local Activation (CLA)}]
\label{def:convexity}
Let the given latent code be $z_0$ and the left/right change points be ${}^lp$ and ${}^rp$ respectively as in Definition \ref{def:cp}. The right slope is defined as ${}^rs=(g_{l:1}^i(z_0+{}^rp\cdot\textbf{e}_d)-g_{l:1}^i(z_0))/{}^rp$ and the left slope is ${}^ls=(g_{l:1}^i(z_0+{}^lp\cdot\textbf{e}_d)-g_{l:1}^i(z_0))/{}^lp$ for a latent dimension $d$. With $C_{i,l}(d,z_0) = ({}^rs-{}^ls)/({}^rp-{}^lp)$, the curvature of local activation for the $i$-th neuron in the $l$-th layer around the given latent code $z_0$ is defined as,
\begin{equation}
    \bar{C}_{i,l}(z_0)=\frac{1}{D_z}\sum_{d=1}^{D_z} C_{i,l}(d,z_0).
\end{equation}
\end{definition}

Although the definitions are constructed in the continuous space, we empirically use a grid search with search bound $R=30$, dividing the search range by 20 for experiments throughout the paper. Details of the hyperparameter setting are provided in Appendix A. 

\begin{figure}[h!]
    \centering
    \includegraphics[width=\columnwidth]{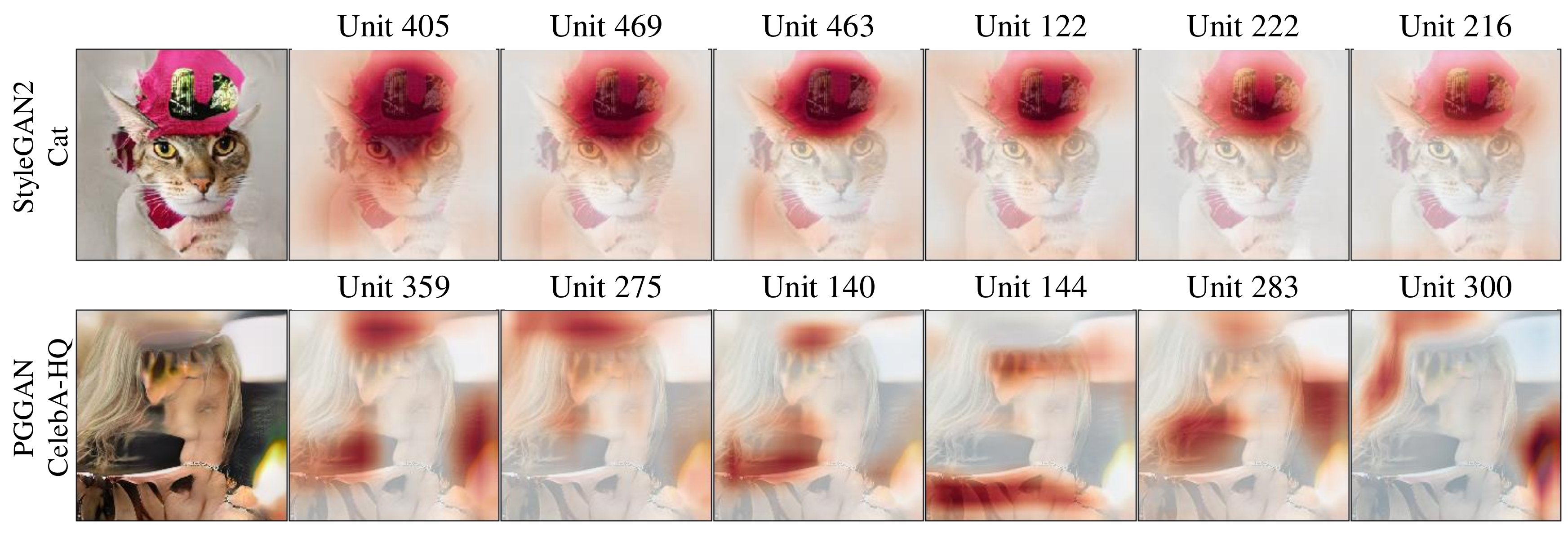}
    \caption{CLA based artifact unit identification. Bi-linear upsampled featuremap units are overlaid with the original generation.}
    \label{fig:unit_identification}
\end{figure}

Figure \ref{fig:unit_identification} shows the featuremap units that have the highest average CLA over the neurons in each unit. We can identify that the activated region for the units with a high CLA is semantically aligned with the artifact area in the generation.

\subsection{Learning Dynamics for Local Activation}
\begin{figure}[h!]
    \centering
    \includegraphics[width=\columnwidth]{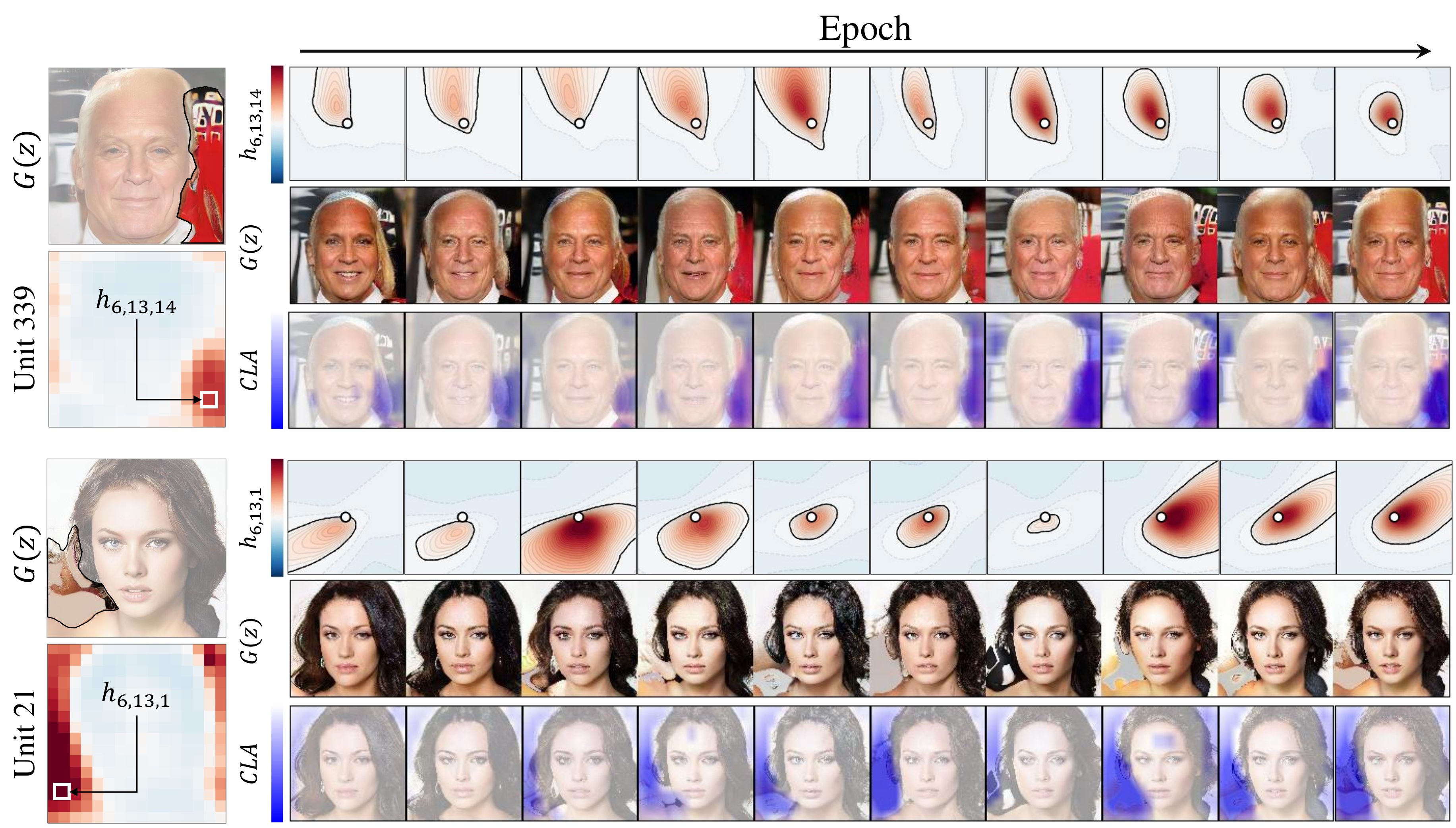}
    \caption{(First row) The visualization of activation pattern in the latent space for the neuron (white box in the featuremap unit) with selected axes. (Second row) The generation $G(z)$ for the fixed latent code. (Third row) Calculated CLA for the neurons in the unit.}
    \label{fig:la_dynamics}
\end{figure}

This section explores the dynamics of CLA for the epochs to validate the correlation between the visual fidelity and the magnitude of CLA. The experiments are performed with pre-trained snapshots of PGGAN model trained on CelebA-HQ\footnote{\label{fn:pggan}\url{https://github.com/tkarras/progressive_growing_of_gans}}. First, we manually select the featuremap unit related to the defective area in layer 6 $\in \mathbb{R}^{512\times16\times16}$. Next, we observe the change of local activation and the artifact emerging process during the training. Figure \ref{fig:la_dynamics} indicates the change of the defective area (low visual fidelity) in the generation and the corresponding CLA. We can identify that the CLA increases when the activation area decreases or the activation value increases in a small area.

\section{Experimental Evaluations}
This section presents analytical results of empirical relations between the low visual fidelity of individual generations and the proposed concept. We select two different GANs with various datasets. We use the pre-trained networks from the authors' official github; (1) PGGAN trained on LSUN-Bedroom, LSUN-Church outdoor \cite{journals/corr/YuZSSX15} and CelebA-HQ$^2$ and (2) StyleGAN2 trained on LSUN-Car, LSUN-Cat, LSUN-Horse \cite{journals/corr/YuZSSX15} and FFHQ\footnote{\url{StyleGAN2:https://github.com/NVlabs/stylegan2}}. To evaluate visual fidelity for individual samples, we define score as,
\begin{align}\nonumber
  S_l(z) &= \sum_i| min(\bar{C}_{i,l}(z),0)*sign(max(h_{l,i},0)) + \\
  &max(\bar{C}_{i,l}(z),0)*sign(min(h_{l,i},0)) |.
\end{align}
The defined score considers the degree of concavity/convexity for positive/negative activation, respectively. If the summation value of the given generation $G(z)$ is larger than other generations, we can expect that $G(z)$ has a low visual fidelity.

\begin{table*}[h!]
\centering\caption{The detection and correction results on various GANs.}\label{tb:1}
\begin{tabular}{c|c|l|cccc}
\hline\hline
Metric & Model & Dataset & Random  & Low CLA   & \multicolumn{1}{c}{High CLA}  & \multicolumn{1}{c}{Correction}\\ \hline

\multirow{7}{*}{RS ($\uparrow$ is better)} &\multirow{3}{*}{PGGAN}     
                          & LSUN-Bedroom & 1.028$\pm$0.003 & \textbf{1.042} & 1.017  & 1.002 \\
                          && LSUN-Church & 1.036$\pm$0.004 & \textbf{1.059} & 1.012    & 1.000 \\
                          && CelebaA-HQ & 1.076$\pm$0.004  & \textbf{1.132} & 1.011   & 1.018 \\ \cline{2-7}
&\multirow{4}{*}{StyleGAN2} & LSUN-Car & 1.066$\pm$0.004 & \textbf{1.084} & 1.044     & 1.061 \\
                          && LSUN-Cat & 1.048$\pm$0.004 & \textbf{1.071} & 1.027    & 1.047 \\
                          && LSUN-Horse &  \textbf{1.056$\pm$0.004}     & 1.046 &  1.053    & 1.054 \\
                          && FFHQ  & 1.075$\pm$0.004 & 1.077 & 1.069      & \textbf{1.097}  \\ \hline\hline
                           
\multirow{7}{*}{PPL ($\downarrow$ is better)} 
&\multirow{3}{*}{PGGAN}     & LSUN-Bedroom   &   423.8$\pm$7.1  &  \textbf{243.9} &  683.3       & -\\
                          && LSUN-Church   &   356.4$\pm$7.9  &  \textbf{213.7} & 558.0       & -\\
                          && CelebaA-HQ    &  243.1$\pm$12.9  & \textbf{114.9} &  443.7        & -\\ \cline{2-7}
&\multirow{4}{*}{StyleGAN2} & LSUN-Car    &  1472.6$\pm$29.3   & \textbf{920.7} &  1938.9        & -\\
                          && LSUN-Cat     &  1501.3$\pm$27.7 & \textbf{1053.2} &  2060.4          & -\\
                          && LSUN-Horse   &  1207.4$\pm$21.5   & \textbf{885.5} &  1552.5        & - \\
                          && FFHQ         &  484.9$\pm$19.2  & \textbf{377.7} &  596.2       & - \\ \hline\hline
\end{tabular}
\end{table*}

\subsection{Qualitative Results}
We randomly select 10k latent codes without truncation for each GAN and calculate the CLA on layer 4 $\in \mathbb{R}^{512\times8\times8}$ for each generation. We choose the top/bottom 1k samples as High/Low CLA groups based on the score, respectively.

\subsubsection{Artifact Detection}
\begin{figure*}[h!]
    \centering
    \includegraphics[width=0.95\textwidth]{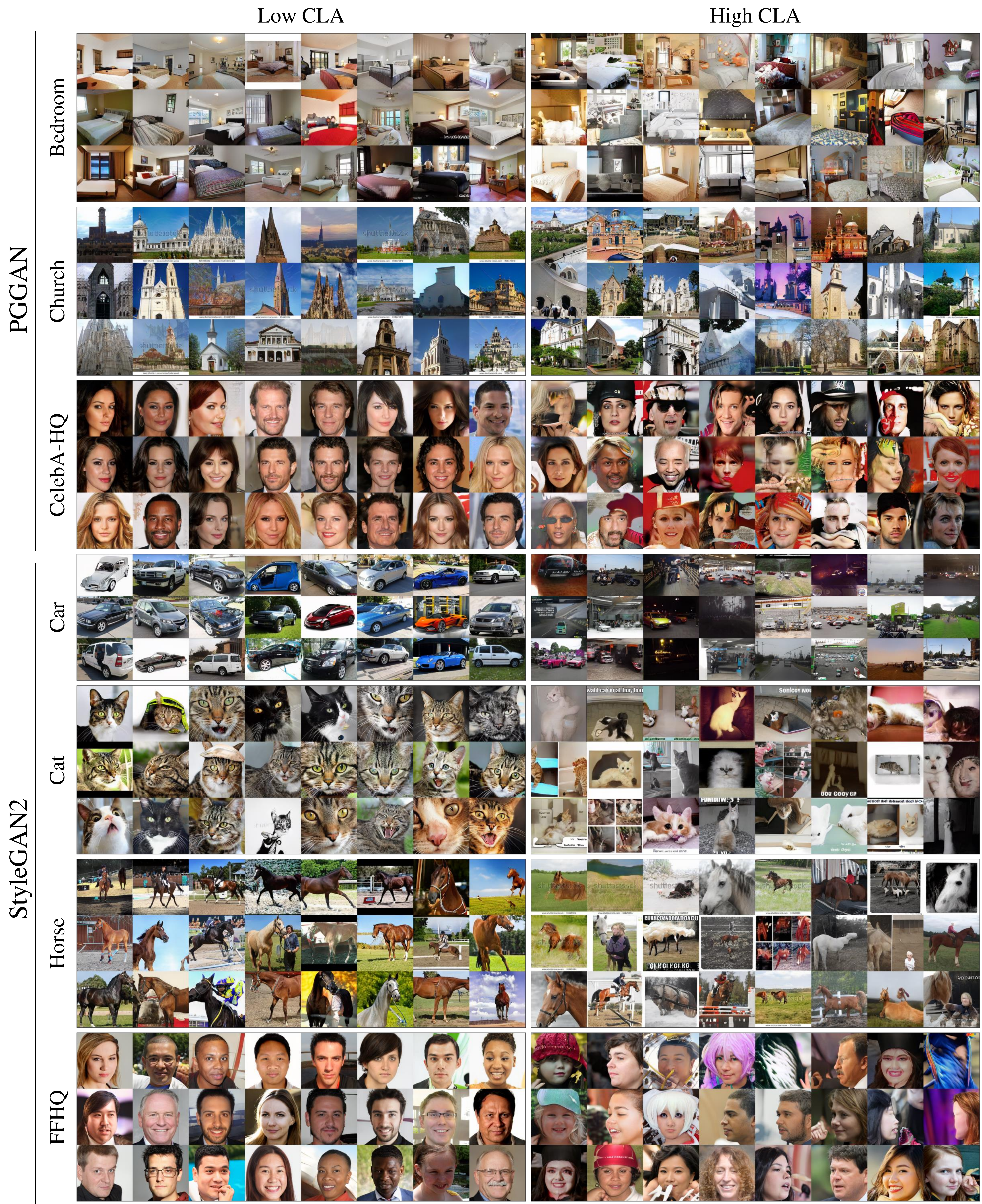}
    \caption{Artifact detection results in various GANs. We select bottom 24 (good) and top 24 (bad) samples for qualitative comparison. We confirm that the generations with high CLA have lower visual fidelity compared to the the generations with low CLA. Appendix C-I provide more examples for artifact detection.}
    \label{fig:qualitative_results}
\end{figure*}

Figure \ref{fig:qualitative_results} depicts the results of detection in each GAN. 
We observe that the generations with a high CLA appear to be more defective than those with a low CLA.
For example, in StyleGAN2 with LSUN-Car, we can identify that the generations which have a high CLA do not include clear information of the car compared with generations that have a low CLA. More detection results are available in the Appendix C-I.

\subsubsection{Artifact Correction}
\begin{figure}[h!]
    \centering
    \includegraphics[width=\columnwidth]{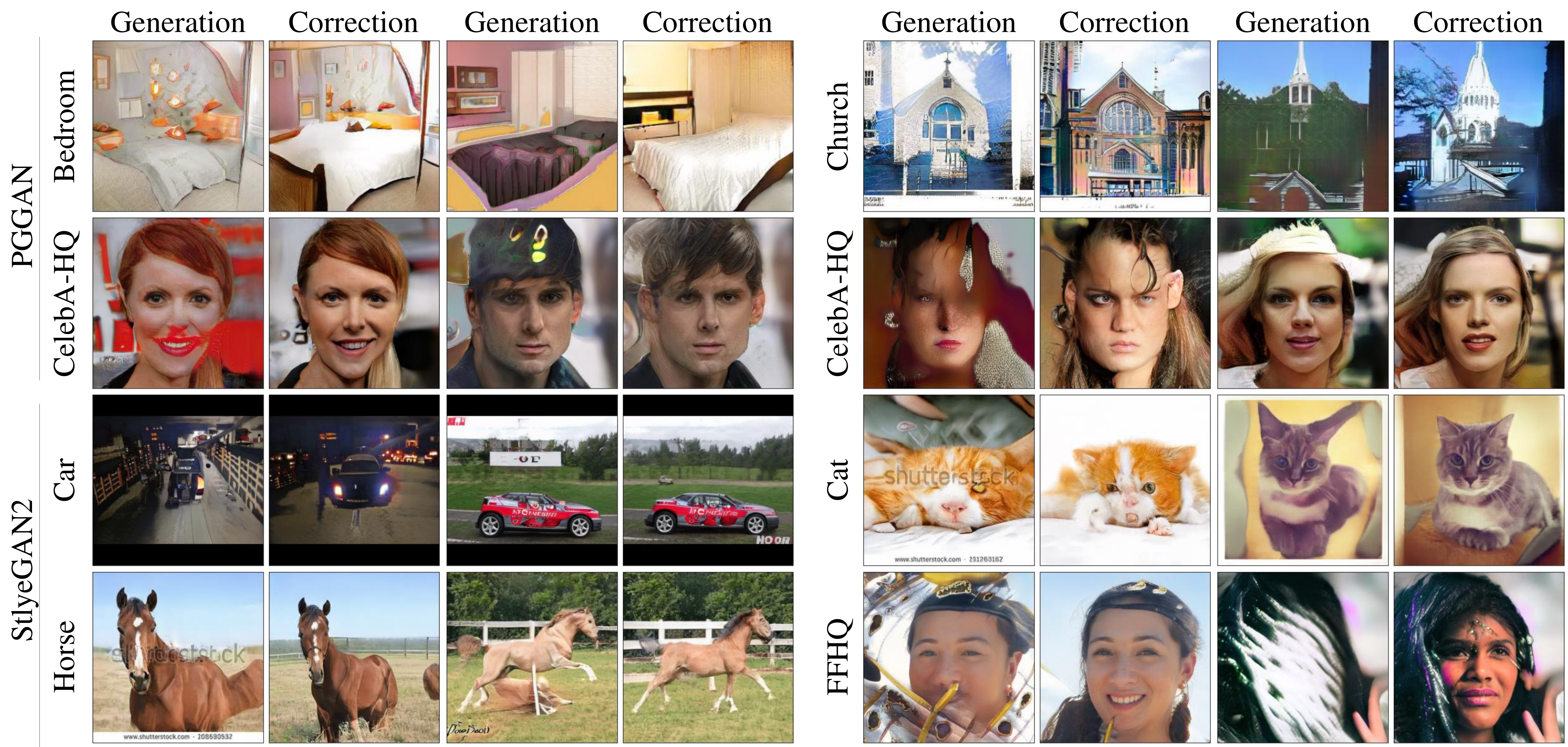}
    \caption{Examples for the correction results on various GANs for High CLA group. More examples are available in Appendix J.}
    \label{fig:corr}
\end{figure}

To validate that the locally activated neurons are related to the artifact, we perform an ablation study on the High CLA group as described in \cite{tousi2021automatic}. Instead of training an external classifier to identify the artifact causing internal units, we use the average CLA over the neurons in each unit. We set the hyperparameters as follows: stopping layer $l=4$, the number of ablation units $n=100$, and the maintain ratio $\lambda=0.9$ for correction. We measure the RS after correction (last column in Table 1.). In Figure \ref{fig:corr}, we observe that when the generations contain severe artifacts, as in the cases of PGGAN with LSUN-Bedroom or LSUN-Church, we may need a more sophisticated method than simple ablation to correct the artifact. Nevertheless, we can improve the visual fidelity in most GANs in the experiments with simple ablation. From the detection and correction experiments, we believe that the locally activated neurons have a strong relationship with low visual-fidelity in the generation.

\subsection{Quantitative Results}\label{sec:quantitative results}
To quantify the fidelity of the detected generations, we calculate RS and PPL for each group; (1) low CLA and (2) high CLA and (3) random selection (30 trials). 
We use 30k real images for each model to calculate RS and set the number of neighborhood $k=3$. For PPL, we perform interpolation in the latent space $z$ with $\epsilon=10^{-4}$. Table \ref{tb:1} indicates the scores for each group in various GANs. We can identify that the high CLA groups have low RS and high PPL compared to random groups. 
The results consistently show that the proposed method can effectively identify the generations with low visual fidelity. 
We note that the proposed method only uses the internal property to evaluate the visual fidelity of individual samples.

\subsubsection{Diversity and Fidelity} To compare diversity and the fidelity in each group, we calculate precision and recall \cite{NEURIPS2019_0234c510} with the truncated samples as the baseline\footnote{The latent $z$ space for PGGAN and the $w$ space for StyleGAN2}.

\begin{figure}[h!]
    \centering
    \includegraphics[width=\columnwidth]{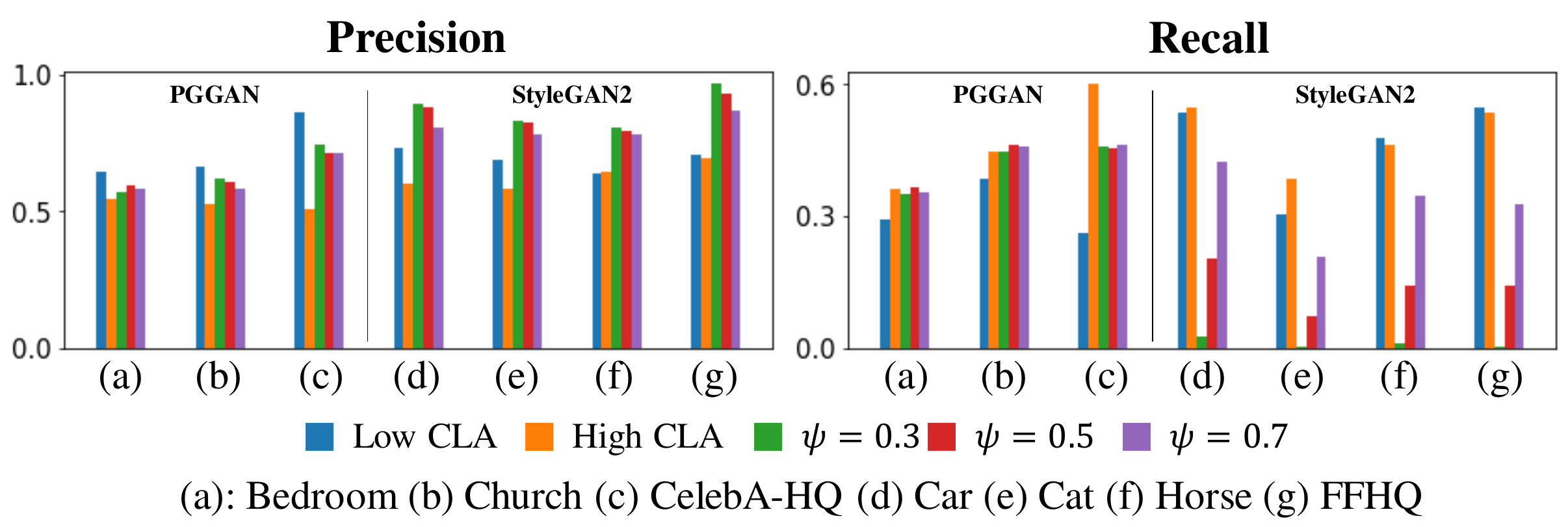}
    \caption{Precision and recall for each group in various GANs.}
    \label{fig:precisionrecall}
\end{figure}

In Figure \ref{fig:precisionrecall}, we can identify that the low CLA group shows the higher precision (fidelity) with slightly lower recall (diversity) comparing to the high CLA group. However, the low CLA group shows much larger recall (diversity) comparing to the truncated samples for StyleGAN2. For PGGAN, the recall is higher on the truncated samples but the precision is higher on the low CLA group.

\section{Discussion}
\subsection{Geometrical Interpretation of LA}\label{section:geo}
In this section, we investigate the relation between the local activation and the visual fidelity under the specific condition.
We begin by specifying the neurons in terms of whether they positively/negatively contribute to the discriminator's decision.
The discriminator $D(x)$ can be similarly described as the generator in Section \ref{sec:quant} where $x$ is the target to generate such as a face image.
The output of discriminator $y$ for the latent code $z$ is represented as $y=D(G(z))$. When we consider one instance $z_0$ and the corresponding feature vector for the $l-1$-th layer $\bar{h}_{l-1}$, one can linearize the networks $G$ and $D$ with a piece-wise linear activation function which is commonly used in the modern GANs. Let $\gamma\ge0$ be the slope parameter for LeakyReLU (in ReLU case, $\gamma$ is zero) and $w_{g_l,i}$ be the $i$-th column vector. The corresponding linearized parameter $\bar{w}_{g_l,i}$ is defined as,
\begin{align}
    \bar{w}_{g_l,i}=
    \begin{cases}
    w_{g_l,i} &\text{where} \quad w_{g_l,i}^\top\bar{h}_{l-1} \geq 0\\
    \gamma \cdot w_{g_l,i} &\text{otherwise}
    \end{cases}
\end{align}
We can then write the linearized generator as $\bar{G}(z_0)=W_{G}^\top \bar{h}_{l}$ and the linearized discriminator as $\bar{D}(\bar{G}(z_0))=W_D^\top\bar{G}(z_0)$ where $W_G^\top=\bar{w}_{g_L}^\top\cdots \bar{w}_{g_{l+1}}^\top \in \mathbb{R}^{D_x\times D_l}$ and $W_D^\top=\bar{w}_{d_L}^\top\cdots \bar{w}_{d_1}^\top \in \mathbb{R}^{1\times D_x}$. The output of the discriminator $\bar{y}$ is paraphrased as,

\begin{align}
    \bar{y}=\bar{D}(\bar{G}(z_0))=W^\top_D W^\top_G \bar{h}_l=\sum_i W_D^\top W_{G,i}^\top \bar{h}_{l,i}
\end{align}
where $W_{G,i}^\top$ is the $i$-th column vector of $W_{G}^\top$ and $\bar{h}_{l,i}$ is the $i$-th element of the vector $\bar{h}_{l}$. We note that $W_D$ is the normal vector of the decision boundary to score the visual quality (real or fake) of the current generation $W_{G}^\top\bar{h}_{l}$. The current generation can be represented as a linear combination of $\{W_{G,i}^\top\}_i$ with the coefficients $\{\bar{h}_{l,i}\}_i$. The contribution of the $i$-th neuron in the $l$-th layer to the discriminator output $\bar{y}$ is $W_D^\top W_{G,i}^\top\bar{h}_{l,i}$. 
We determine that the $i$-th neuron has a negative/positive contribution if the contribution of the $i$-th neuron decreases/increases for the decision of the discriminator. For example, when $W_D^\top W_{G,i}^\top\bar{h}_{l,i} <0$, the $i$-th neuron has a negative contribution.

\begin{figure}[h!]
    \centering
    \includegraphics[width=\columnwidth]{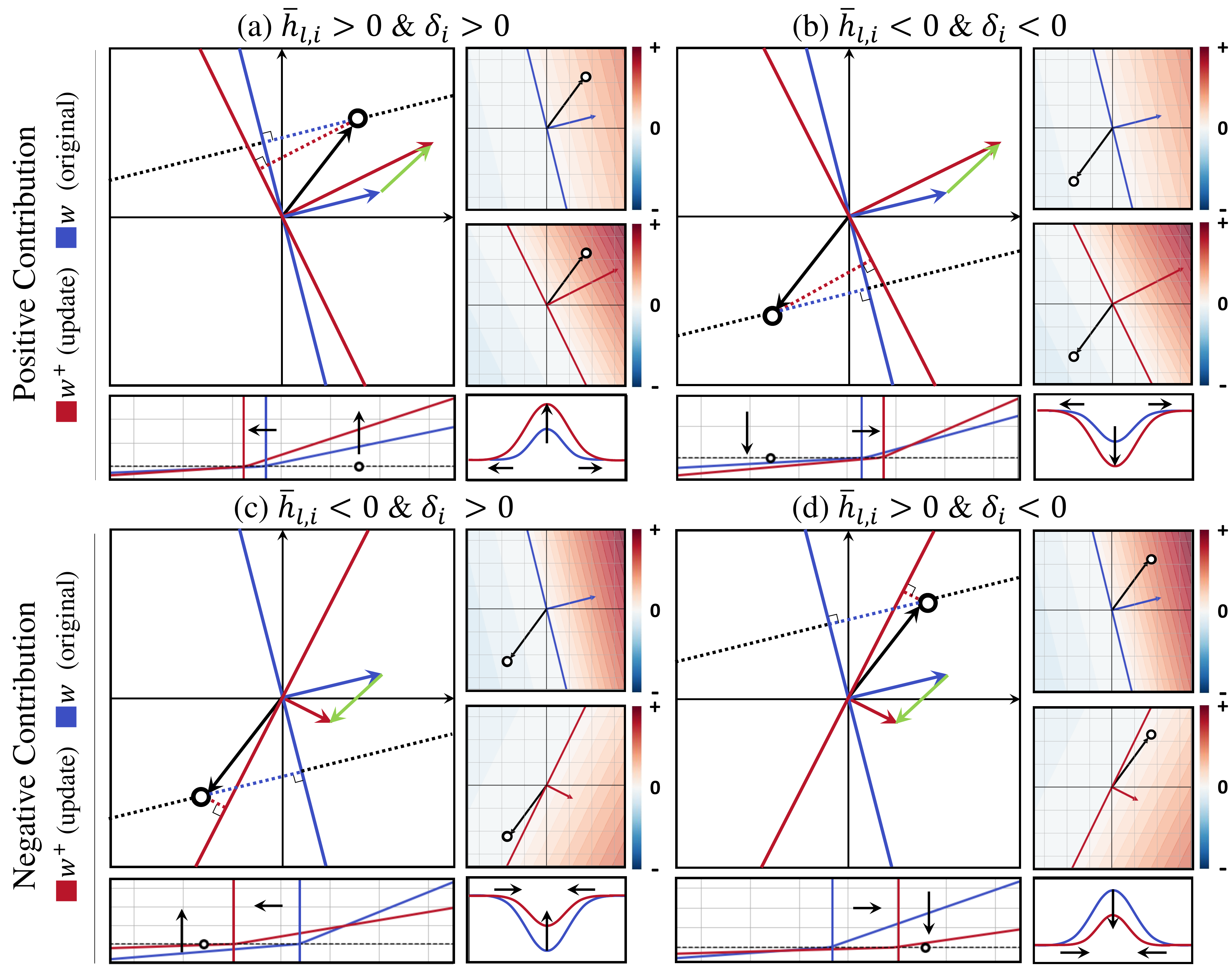}
    \caption{The geometrical illustrations of update cases ($\bar{h}_{l-1} \rightarrow \bar{h}_{l,i}$). The black arrow and the dot represent $\bar{h}_{l-1}$, the green arrow is $\delta_i \bar{h}_{l-1}$ (update term), the blue arrow is $\bar{w}_{g_l,i}$ (original parameter) and the red arrow is $\bar{w}_{g_l,i}^+$ (updated parameter). The bottom-left plot indicates the activation values on the black dashed line before and after the update. The bottom-right plot shows the conceptual representation of the activation pattern change after the update ($z_0 \rightarrow \bar{h}_{l,i}$).  
    }
    \label{fig:update_examples}
\end{figure}

We perform a geometrical analysis for the neurons that have a negative/positive contribution to the output of discriminator in the vanilla GAN \cite{goodfellow2014generative}. We begin with describing how to update the parameters related to the direction of the contribution of each neuron and then analyze the consequences of updates. The loss function for the generator $G$ and discriminator $D$ is defined as,
\begin{align}\label{eq:loss}\nonumber
  \min\limits_{G}\max\limits_{D}V(D,G)=E_{x\sim p_{data}(x)}[\log f(D(x))]+\\
  E_{x\sim p_{z}(z)}[\log(1-f(D(G(z))))]  
\end{align}
where $f(\cdot)$ is the sigmoid function. For $w_{g_l}$, the updated parameter $w_{g_l}^+$ by the stochastic gradient descent of the given latent code $z_0$ with linearized form is described as,
\begin{equation}
    w_{g_l}^+ = w_{g_l} + \eta c_0 f'(\bar{y})W_D^\top W_G^\top\bar{h}_{l-1}
    \label{eq:linear_gradient}
\end{equation}

where $c_0 = (1-f(D(G(z_0))))^{-1}$ 
and $\eta$ is learning rate. The $i$-th column vector $w_{g_l,i}$ induces activation of the $i$-th neuron in the $l$-th layer ($\bar{h}_{l,i}$), and is updated by the direction $\bar{h}_{l-1}$ with weight $\delta_i=\eta c_0 f'(\bar{y})W_D^\top  W_{G,i}^\top \in \mathbb{R}$.
Figure \ref{fig:update_examples} presents geometrical illustrations of the update for four possible cases.

In Figure \ref{fig:update_examples}, we can observe that the perpendicular distance between the generative boundary and the feature vector $\bar{h}_{l,i}$ (colored dotted lines) decreases in the negative contribution cases when the learning rate is sufficiently small. Further, the magnitude of the activation value for $\bar{h}_{l,i}$ also decreases in the negative contribution cases ($\bar{h}_{l,i}\cdot\delta_i<0$) as 
\begin{align}\nonumber
    \bar{h}_{l-1}^{\top}w_{g_l,i}^+ &= \bar{h}_{l-1}^{\top}(w_{g_l,i} + \delta_i\bar{h}_{l-1}) \\
    &= \bar{h}_{l-1}^{\top}w_{g_l,i} + \delta_i\Vert\bar{h}_{l-1}\Vert^2.
\end{align}
From the analysis on the vanilla GAN, we can presume that when a neuron in the generator negatively contributes to the discriminator output, the generator tries to deactivate the neuron by reducing the activation and distance to deceive the discriminator.
As the penalization can be applied for the arbitrary latent code $z$, if a neuron has the negative contribution consistently during the training, the corresponding activated region in the latent space will shrink.
If the locally activated region is not fully removed during the training, the corresponding neuron may generate artifacts when highly activated.

We note that although the analysis can suggest the partial explanations, the theoretical reasons for the observed relation are still remained as an open question.

\subsection{Comparisons with PPL}
We discuss the differences from the Perceptual Path Length (PPL), which is the most similar to the proposed method among feasibility metrics in that both measure the smoothness of the latent space of the generator.
The first difference is that PPL needs an external network (e.g., pre-trained VGG16) to quantify the smoothness in the latent space. The dependency on the external network not only requires additional resources but also raises a limitation that the reliability depends on the capability of the external network. 
In other words, if the class of generations is not well-aligned to the external network, it becomes non-trivial to guarantee the reliability of the quantified perceptual distance in the feature space of the external network. 

Secondly, PPL measures the first order derivative whereas our method measures the second order derivative. Even though the path length regularization is applied in StyleGAN v2 to regularize the first order derivative, we can still observe high CLA.

\section{Conclusion}
In this paper, we propose the concept \textit{local activation} on the internal neurons of GANs to evaluate the low visual fidelity of generations. 
We further discuss an analysis on the relationship between the proposed concept and low visual fidelity of individual generations under the restricted condition. 
We perform empirical studies to validate the relations on artifact detection and correction settings. The proposed method shows reasonable performance without additional supervision or resources. 
Because the proposed method uses the basic element (neuron) of neural networks and its internal information, we believe that the proposed approach can be extended to a wide range of deep neural networks.

\newpage
\section*{Acknowledgements} This work was conducted by Center for Applied Research in Artificial Intelligence (CARAI) grant funded by DAPA and ADD (UD190031RD), Institute of Information \& communications Technology Planning \& Evaluation (IITP) grant funded by the Korea government (MSIT) (No.2017-0-01779, XAI and No.2019-0-00075, Artificial Intelligence Graduate School Program (KAIST)) and partly supported by KAIST-NAVER Hypercreative AI Center.

\bibliography{aaai22}

\appendix
\onecolumn
\renewcommand{\thesection}{\Alph{section}}

\title{Appendix: \\ An Unsupervised Way to Understand Artifact Generating Internal Units \\in Generative Neural Networks}
\maketitle

\section{Hyperparameter Exploration}\label{sec:hyperparam}
\begin{figure}[h!]
    \centering
    \includegraphics[width=0.6\columnwidth]{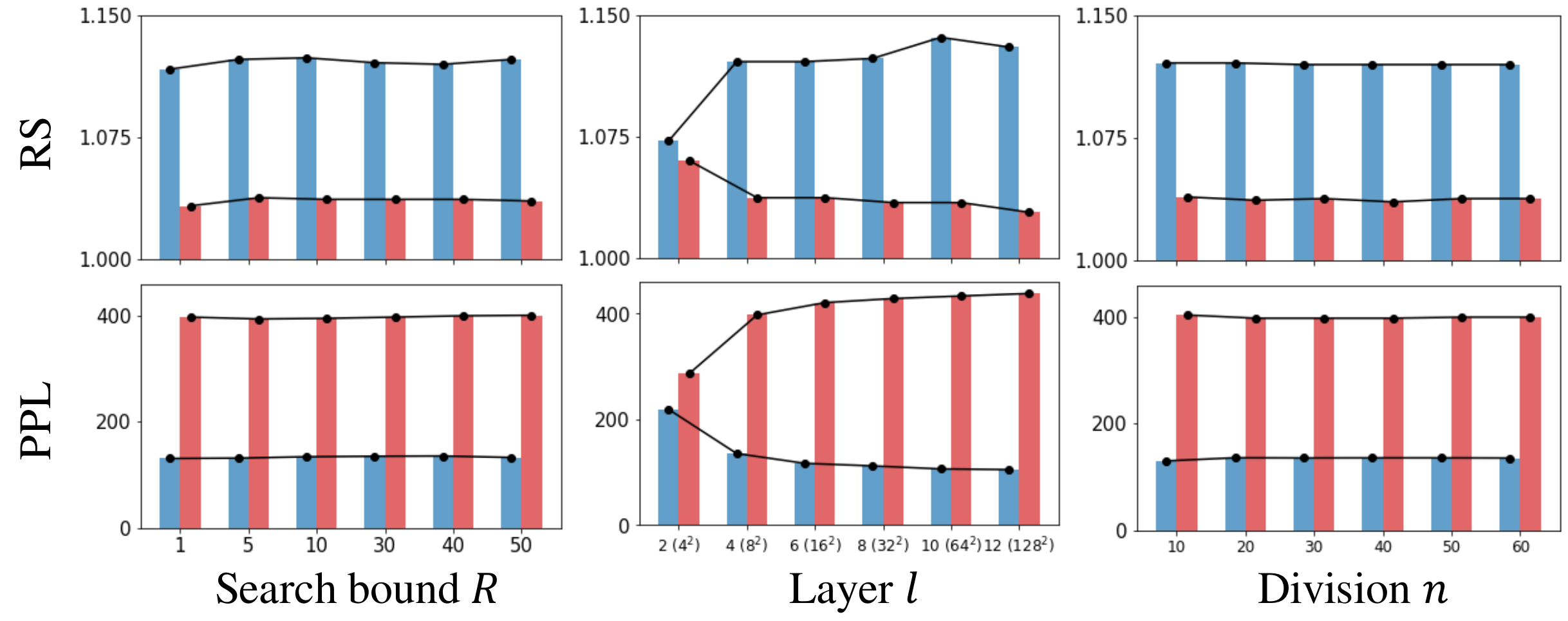}
    \caption{RS and PPL scores for the various hyperparameter settings. (First column) $n=20$ and $l=4$ are fixed. (Second column) $R=30$ and $n=20$ are fixed. (Third column) $R=30$ and $l=4$ are fixed.}\label{fig:hyp_exp}
\end{figure}

We perform the exploration to set the hyperparameters such as search bound $R$, division $n$, and target layer $l$. We select 200 samples for each group (low/high CLA) in random 1k generations for PGGAN with CelebA-HQ and calculate the average of RS and PPL of each group for various hyperparameter settings. In Figure \ref{fig:hyp_exp}, the blue/red color means low/high CLA group respectively.

From first and third columns, it is hard to identify the big difference of detection performance for the magnitude of $R$. To apply to various GANs, we prefer the larger value of $R$ to guarantee the existence of the CPs which make the CLA computation more accurate. However, the larger value of $R$ may require the larger value of division $n$ which requires a high computational cost. Considering the trade-off we set $R=30$ and $n=20$. In the second column, we can identify that the detection performance increases depending on the depth of layer, although the increase is small enough after layer 4. Because the number of inspection neurons increases exponentially which means that the computation time increases exponentially, we select layer 4 as the target layer for the detection experiment as considering both performances of detection and computation time.

\section{Computational Complexity}
The computational complexity is roughly approximated as 
$$O(|D_z|*n) + O(|D_l|*|D_z|*n)$$ 
where the first term is for the forward pass and second term is for the calculation of CLA. We use 1 GPU (RTX 2080 Ti) to measure the computation time with 100 iterations. In PGGAN, the calculation time for one latent code is about 5.6 $\pm$ 0.31 sec  at the target layer (512x8x8). In StyleGAN2, the calculation time for the one latent code is about 26.4 $\pm$ 0.41 sec at the target layer (512x8x8).

The computation can be implemented by the parallel computing (multi GPUs) to boost the speed, because we can parallelize the computation for each latent dimension. For example, if we use 8 GPUs to divide axes, the total calculation time decreases by 1/8. We believe that the computation can be further improved by optimizing the code.

\newpage
\section{Artifact Detection in StyleGAN2-LSUN Car}\label{sec:detection_car}
\begin{figure}[h!]
    \centering
     \begin{subfigure}[b]{0.9\columnwidth}
         \centering
         \includegraphics[width=\textwidth]{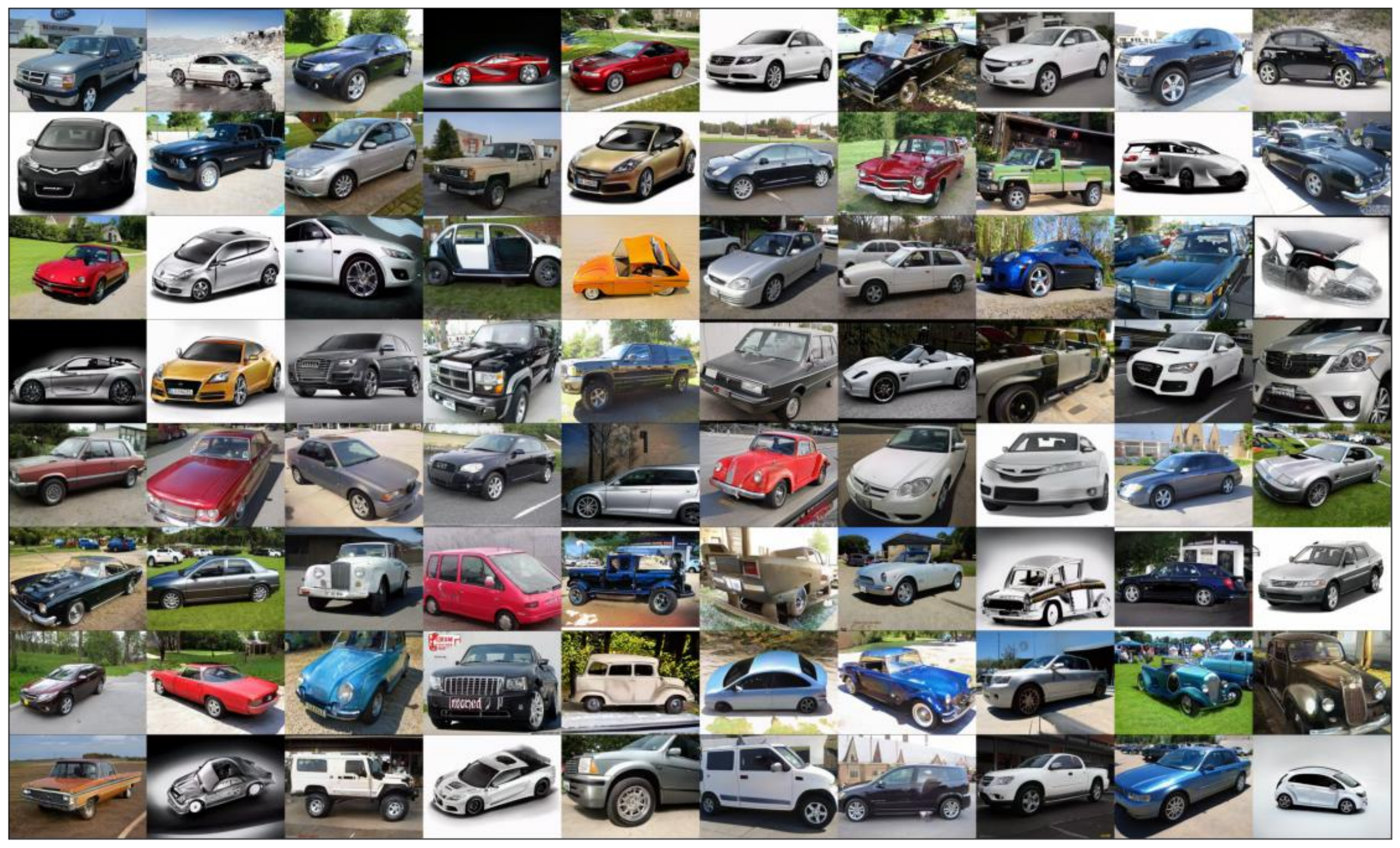}
         \caption{Bottom 25-104 samples in low CLA group.}
     \end{subfigure}
    \begin{subfigure}[b]{0.9\columnwidth}
         \centering
         \includegraphics[width=\textwidth]{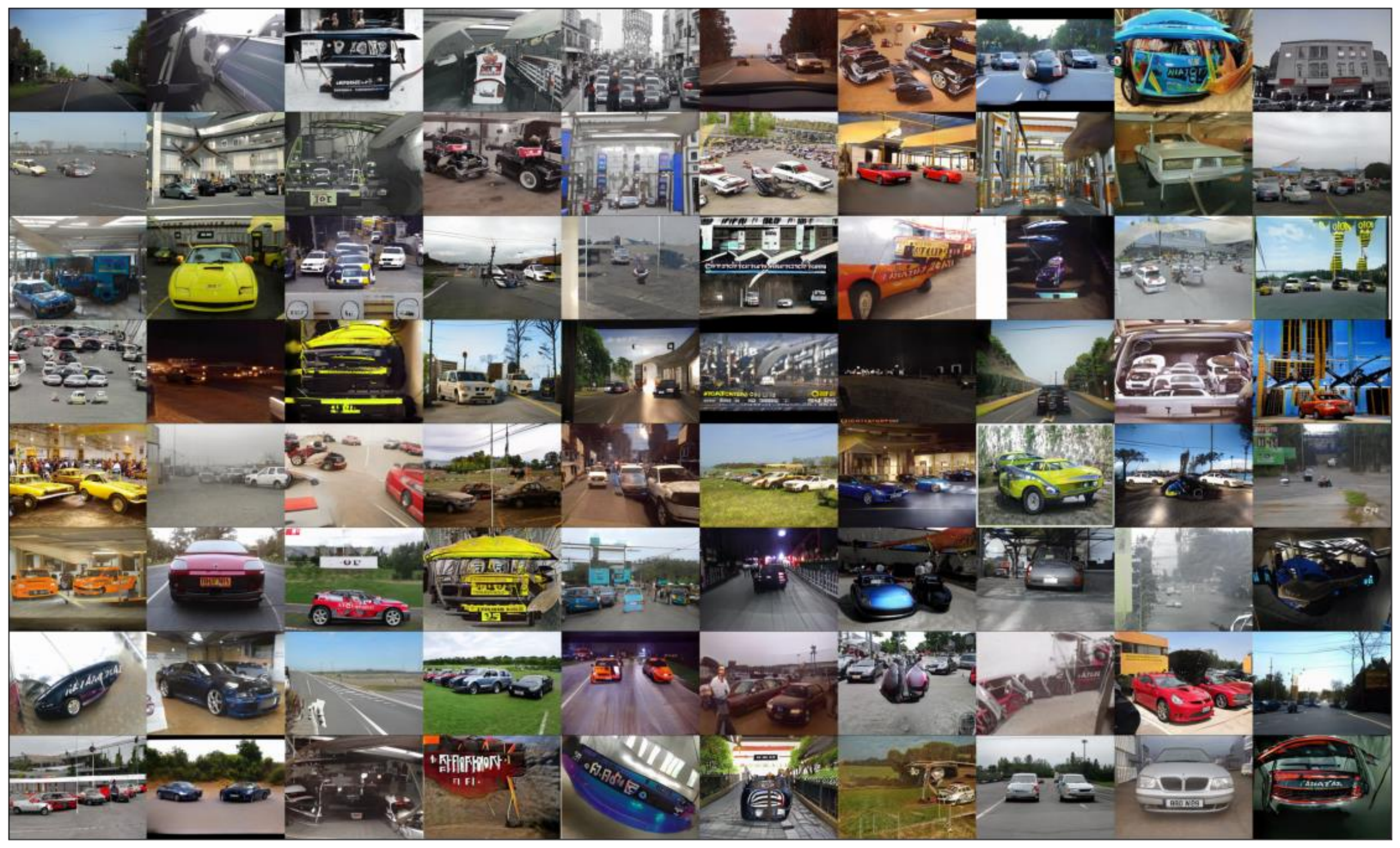}
         \caption{Top 25-104 samples in High CLA group.}
     \end{subfigure}

     \caption{Samples in the order of CLA in StyleGAN2 with LSUN-Car.}
     \label{fig:stylegan_car}
\end{figure}

\newpage
\section{Artifact Detection in StyleGAN2-LSUN Cat}
\begin{figure}[h!]
    \centering
     \begin{subfigure}[b]{0.9\columnwidth}
         \centering
         \includegraphics[width=\textwidth]{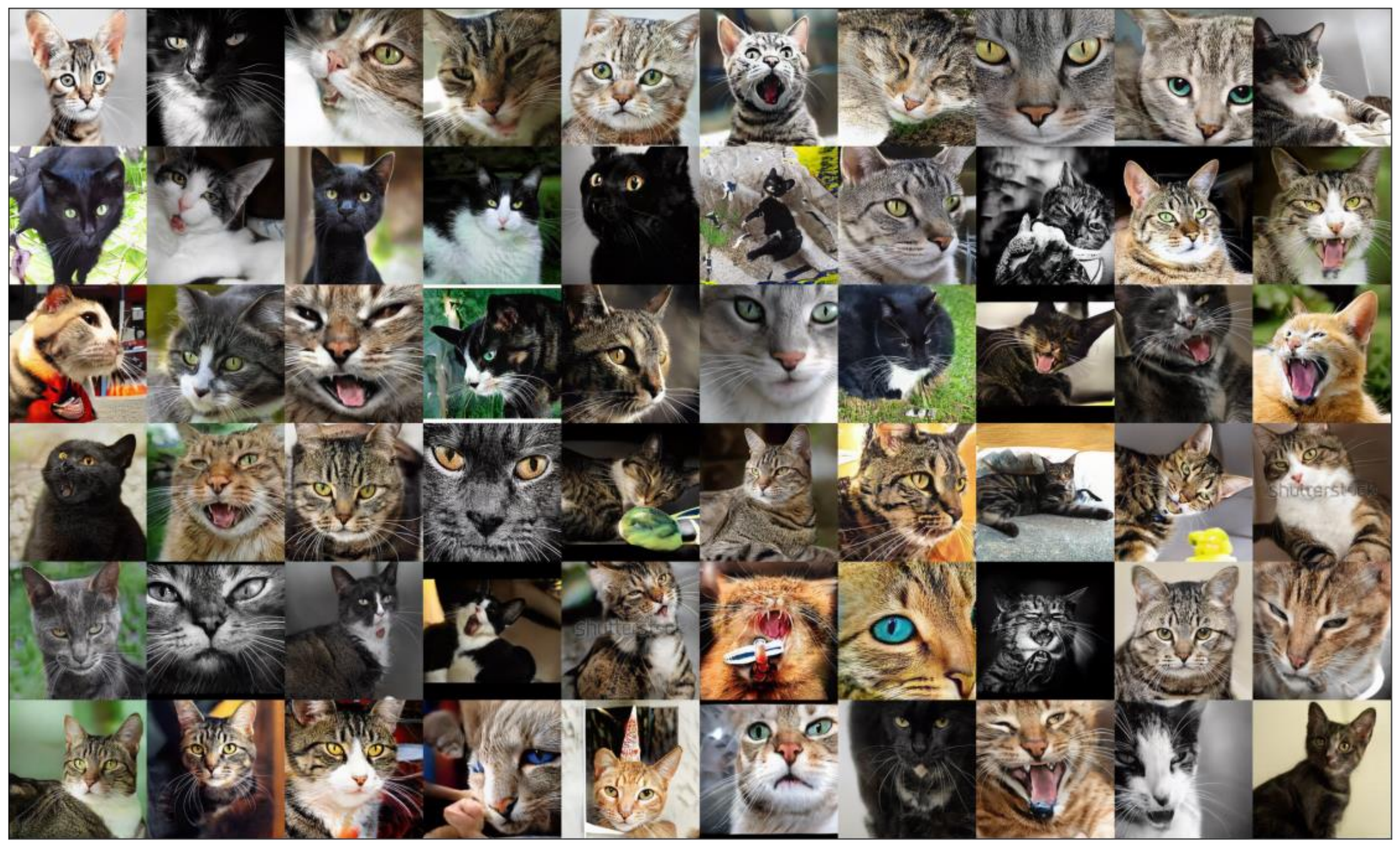}
         \caption{Bottom 25-84 samples in low CLA group.}
     \end{subfigure}
    \begin{subfigure}[b]{0.9\columnwidth}
         \centering
         \includegraphics[width=\textwidth]{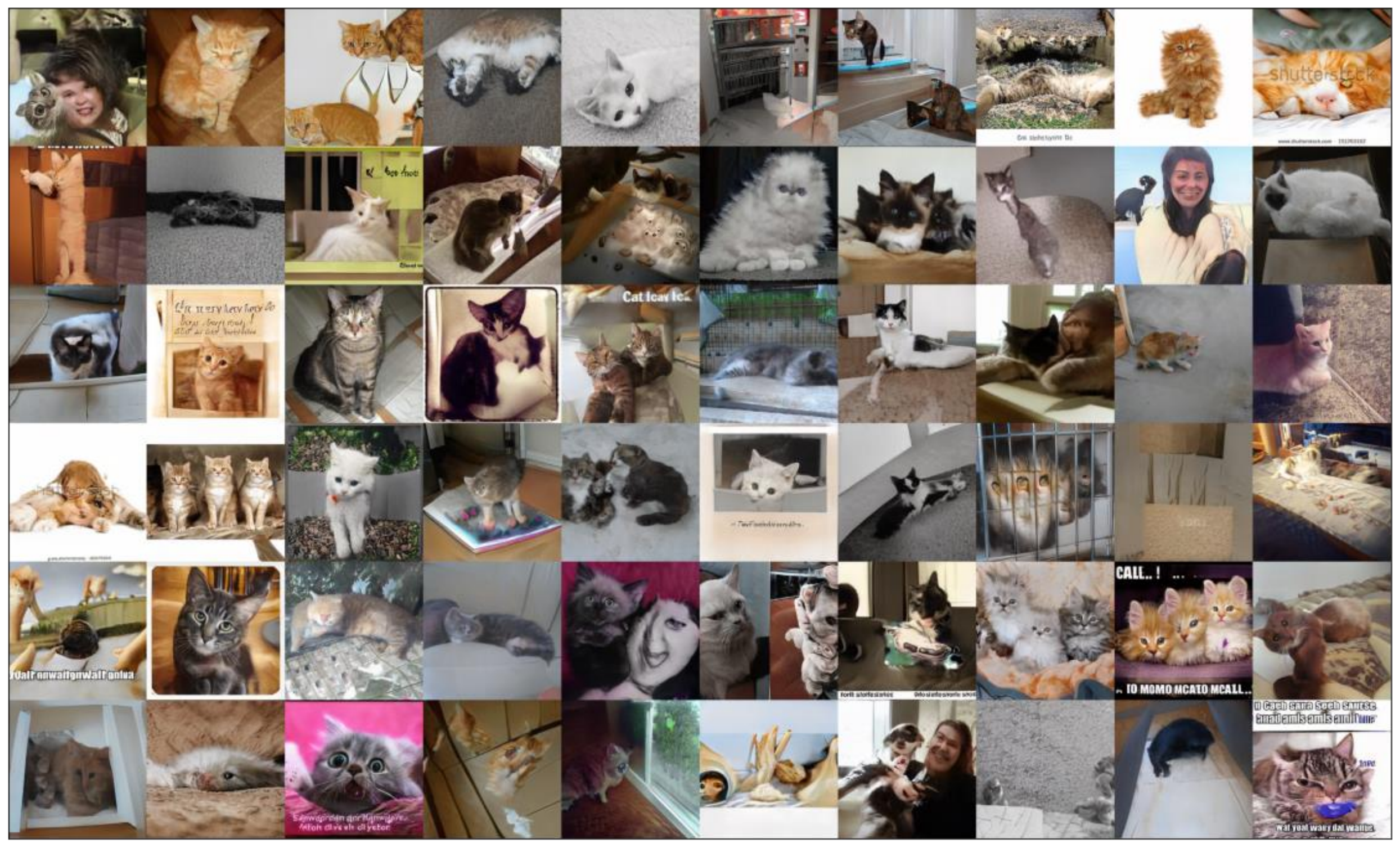}
         \caption{Top 25-84 samples in High CLA group.}
     \end{subfigure}

     \caption{Samples in the order of CLA in StyleGAN2 with LSUN-Cat.}
     \label{fig:stylegan_cat}
\end{figure}

\newpage
\section{Artifact Detection in StyleGAN2-LSUN Horse}
\begin{figure}[h!]
    \centering
     \begin{subfigure}[b]{0.9\columnwidth}
         \centering
         \includegraphics[width=\textwidth]{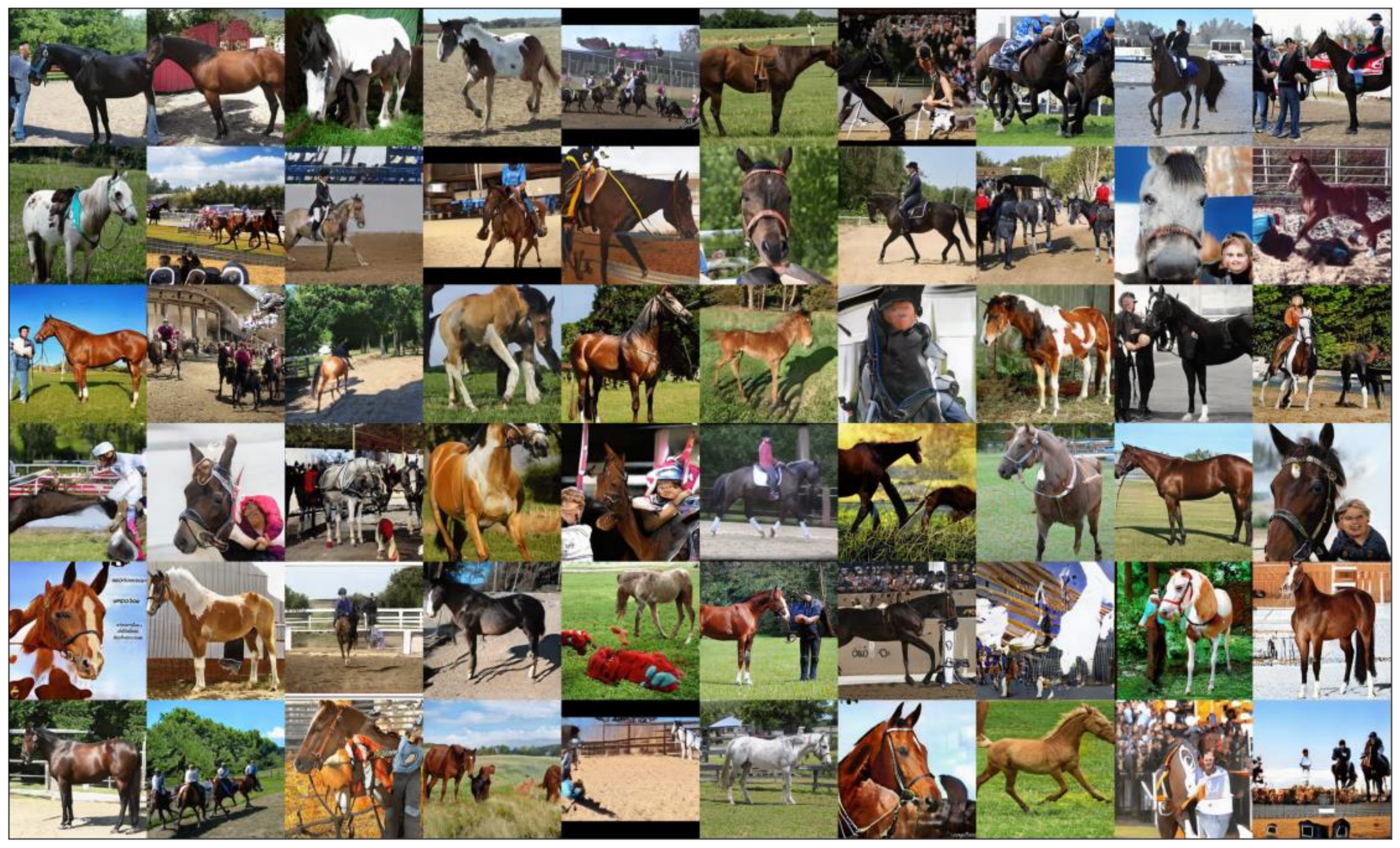}
         \caption{Bottom 25-84 samples in low CLA group.}
     \end{subfigure}
    \begin{subfigure}[b]{0.9\columnwidth}
         \centering
         \includegraphics[width=\textwidth]{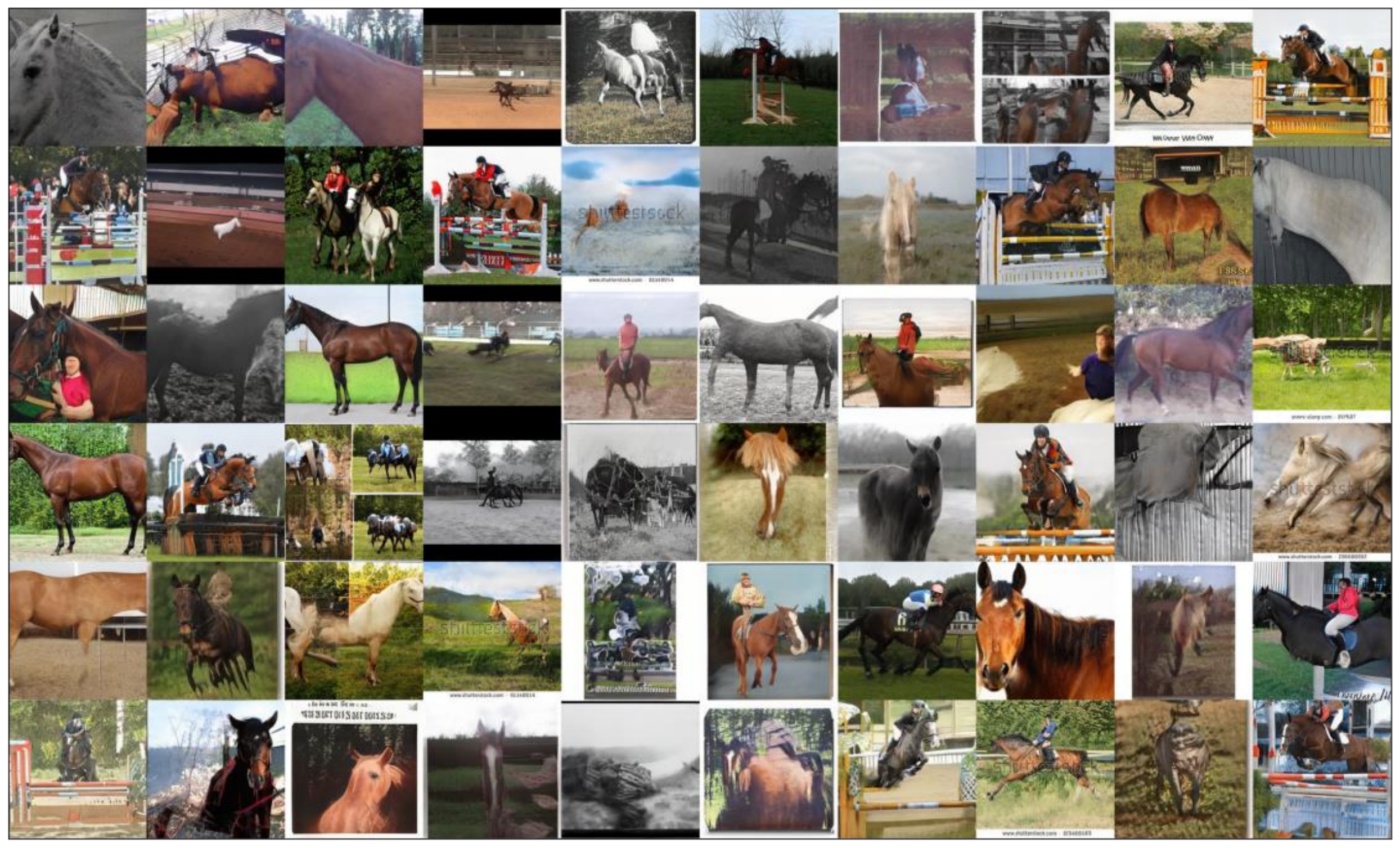}
         \caption{Top 25-84 samples in High CLA group.}
     \end{subfigure}

     \caption{Samples in the order of CLA in StyleGAN2 with LSUN-Horse.}
     \label{fig:stylegan_horse}
\end{figure}

\newpage
\section{Artifact Detection in StyleGAN2-FFHQ}
\begin{figure}[h!]
    \centering
     \begin{subfigure}[b]{0.9\columnwidth}
         \centering
         \includegraphics[width=\textwidth]{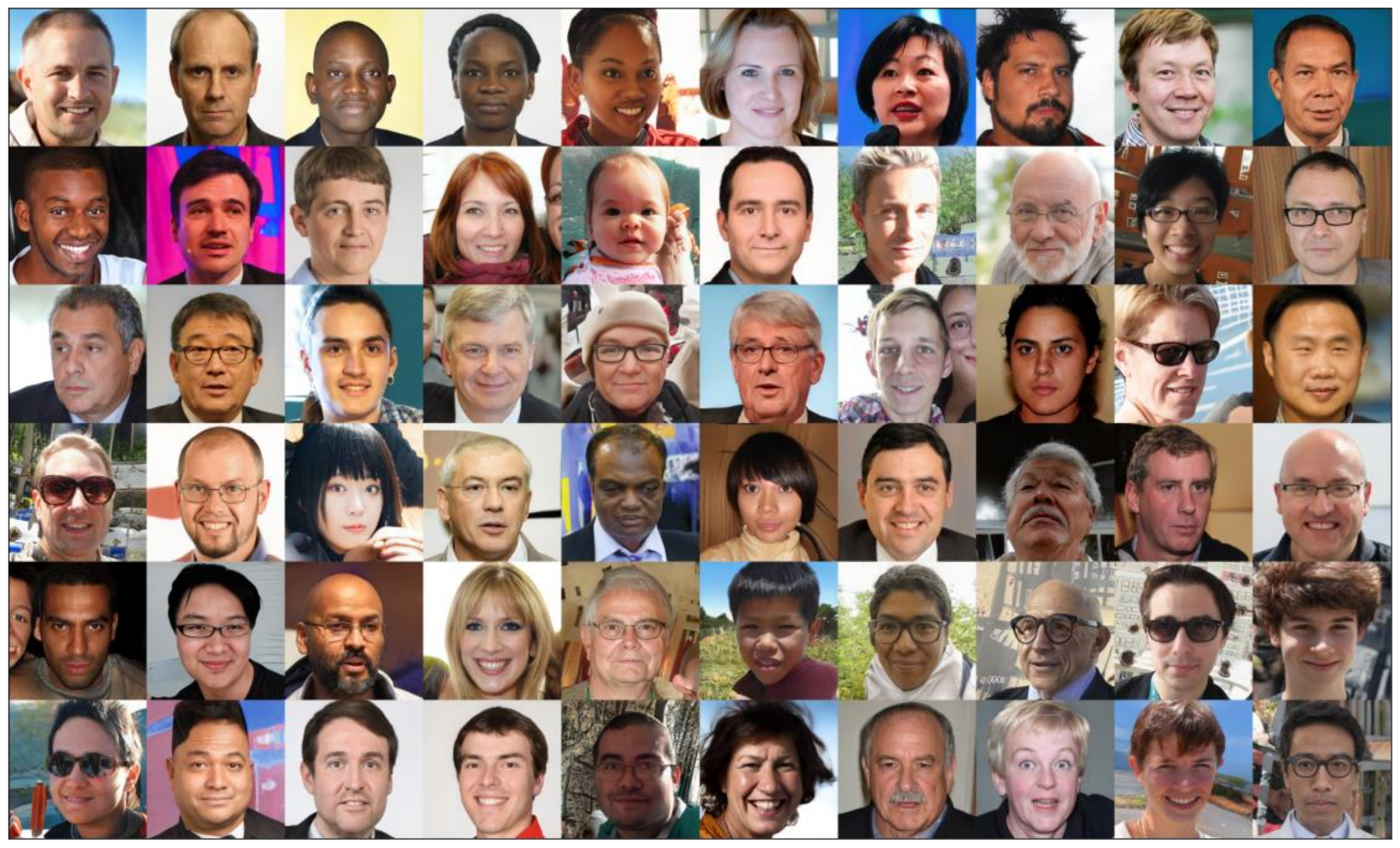}
         \caption{Bottom 25-84 samples in low CLA group.}
     \end{subfigure}
    \begin{subfigure}[b]{0.9\columnwidth}
         \centering
         \includegraphics[width=\textwidth]{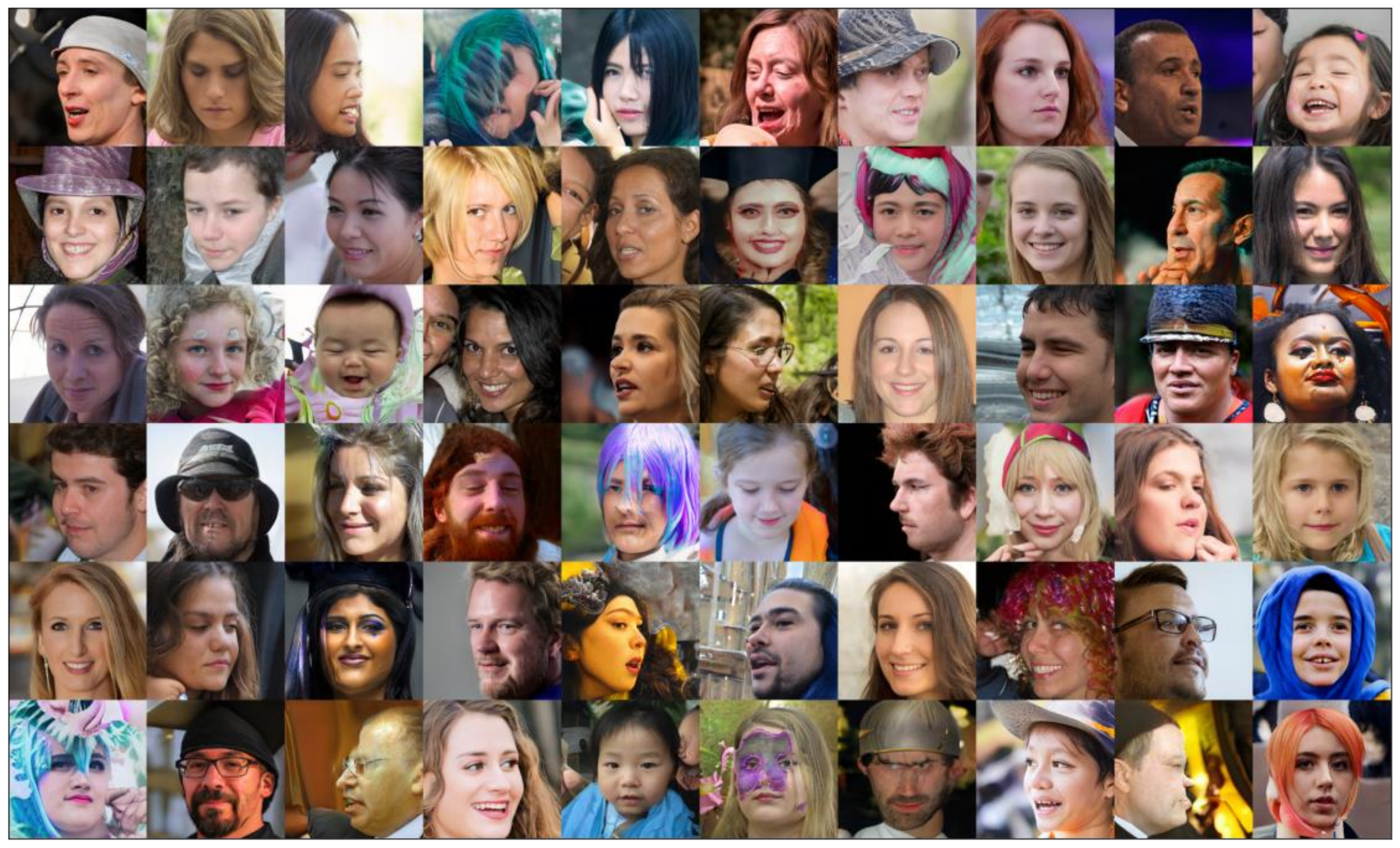}
         \caption{Top 25-84 samples in High CLA group.}
     \end{subfigure}

     \caption{Samples in the order of CLA in StyleGAN2 with FFHQ.}
     \label{fig:stylegan_ffhq}
\end{figure}

\newpage
\section{Artifact Detection in PGGAN-LSUN Bedroom}
\begin{figure}[h!]
    \centering
     \begin{subfigure}[b]{0.9\columnwidth}
         \centering
         \includegraphics[width=\textwidth]{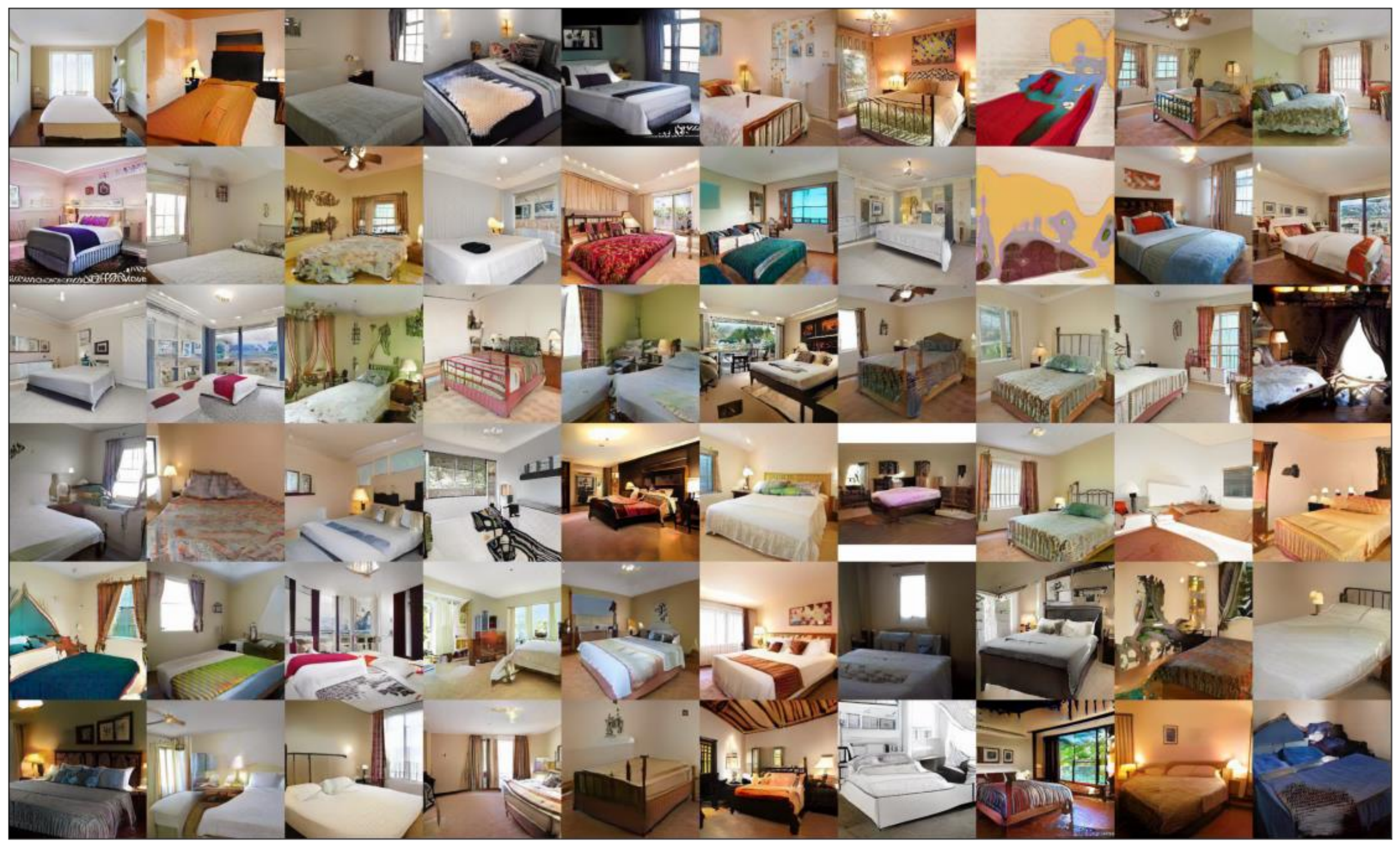}
         \caption{Bottom 25-84 samples in low CLA group.}
     \end{subfigure}
    \begin{subfigure}[b]{0.9\columnwidth}
         \centering
         \includegraphics[width=\textwidth]{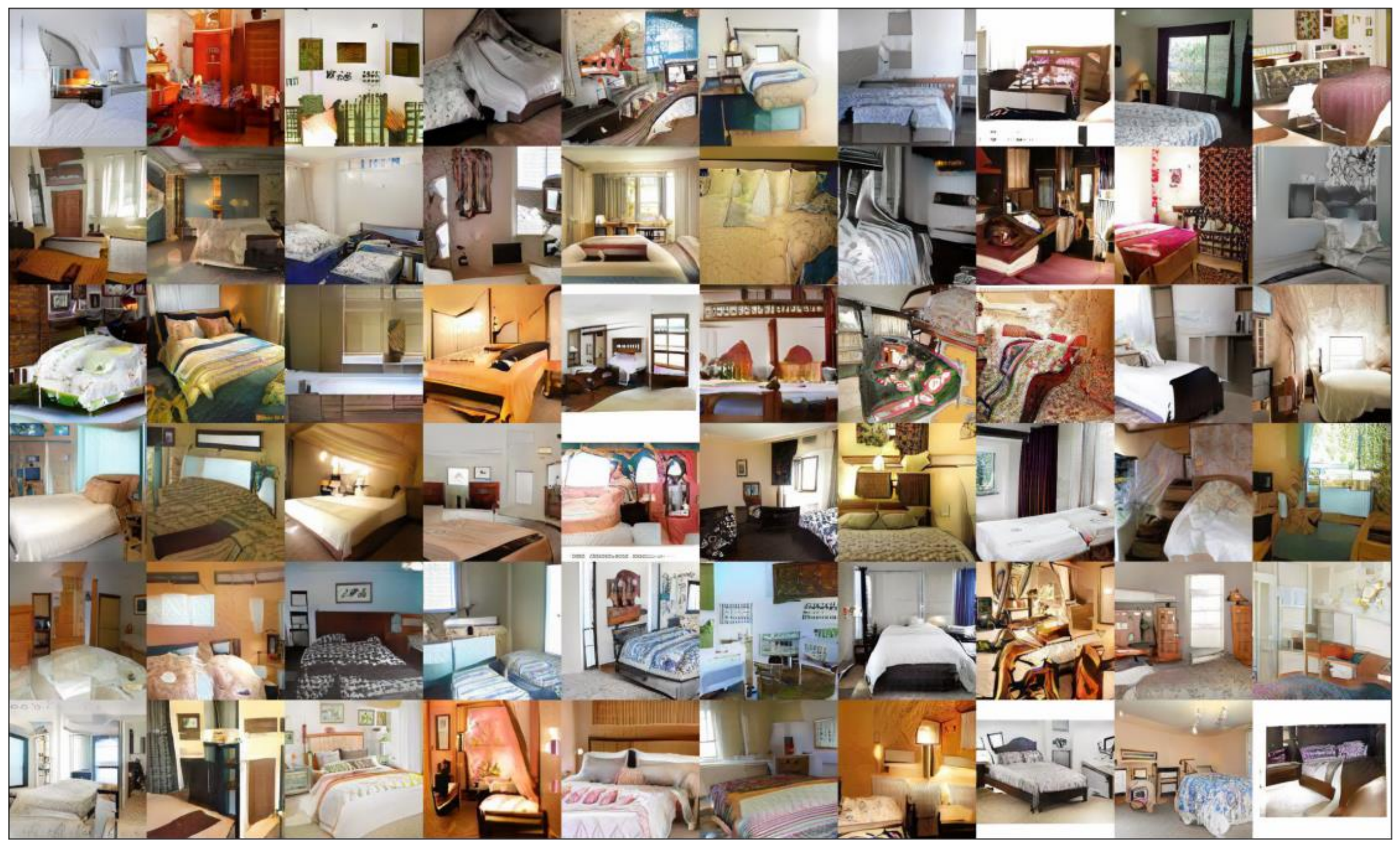}
         \caption{Top 25-84 samples in High CLA group.}
     \end{subfigure}

     \caption{Samples in the order of CLA in PGGAN with LSUN-Bedroom.}
     \label{fig:pggan_bedroom}
\end{figure}

\newpage
\section{Artifact Detection in PGGAN-LSUN Church}
\begin{figure}[h!]
    \centering
     \begin{subfigure}[b]{0.9\columnwidth}
         \centering
         \includegraphics[width=\textwidth]{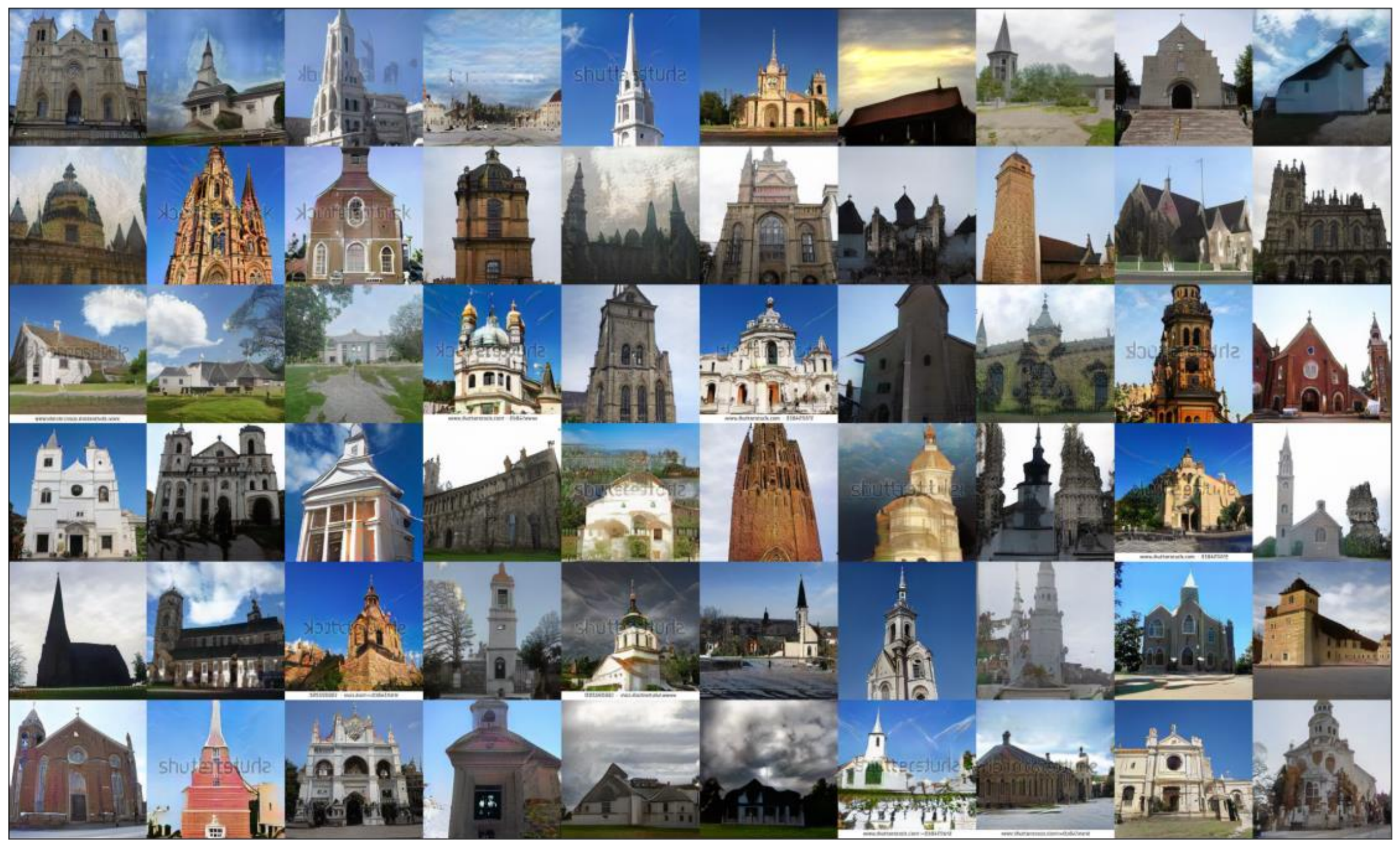}
         \caption{Bottom 25-84 samples in low CLA group.}
     \end{subfigure}
    \begin{subfigure}[b]{0.9\columnwidth}
         \centering
         \includegraphics[width=\textwidth]{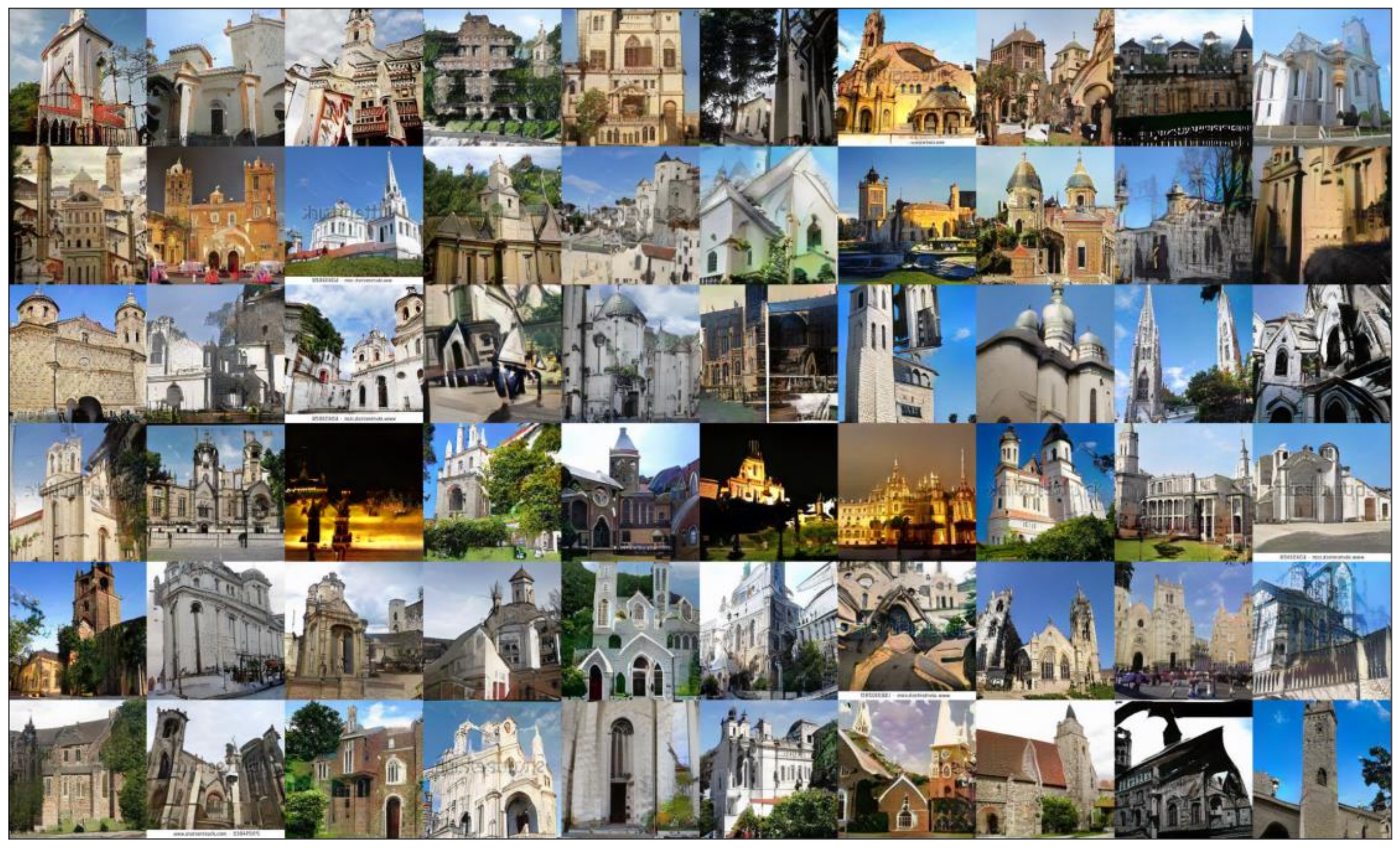}
         \caption{Top 25-84 samples in High CLA group.}
     \end{subfigure}

     \caption{Samples in the order of CLA in PGGAN with LSUN-Church.}
     \label{fig:pggan_church}
\end{figure}

\newpage
\section{Artifact Detection in PGGAN-CelebA HQ}\label{sec:correction}
\begin{figure}[h!]
    \centering
     \begin{subfigure}[b]{0.9\columnwidth}
         \centering
         \includegraphics[width=\textwidth]{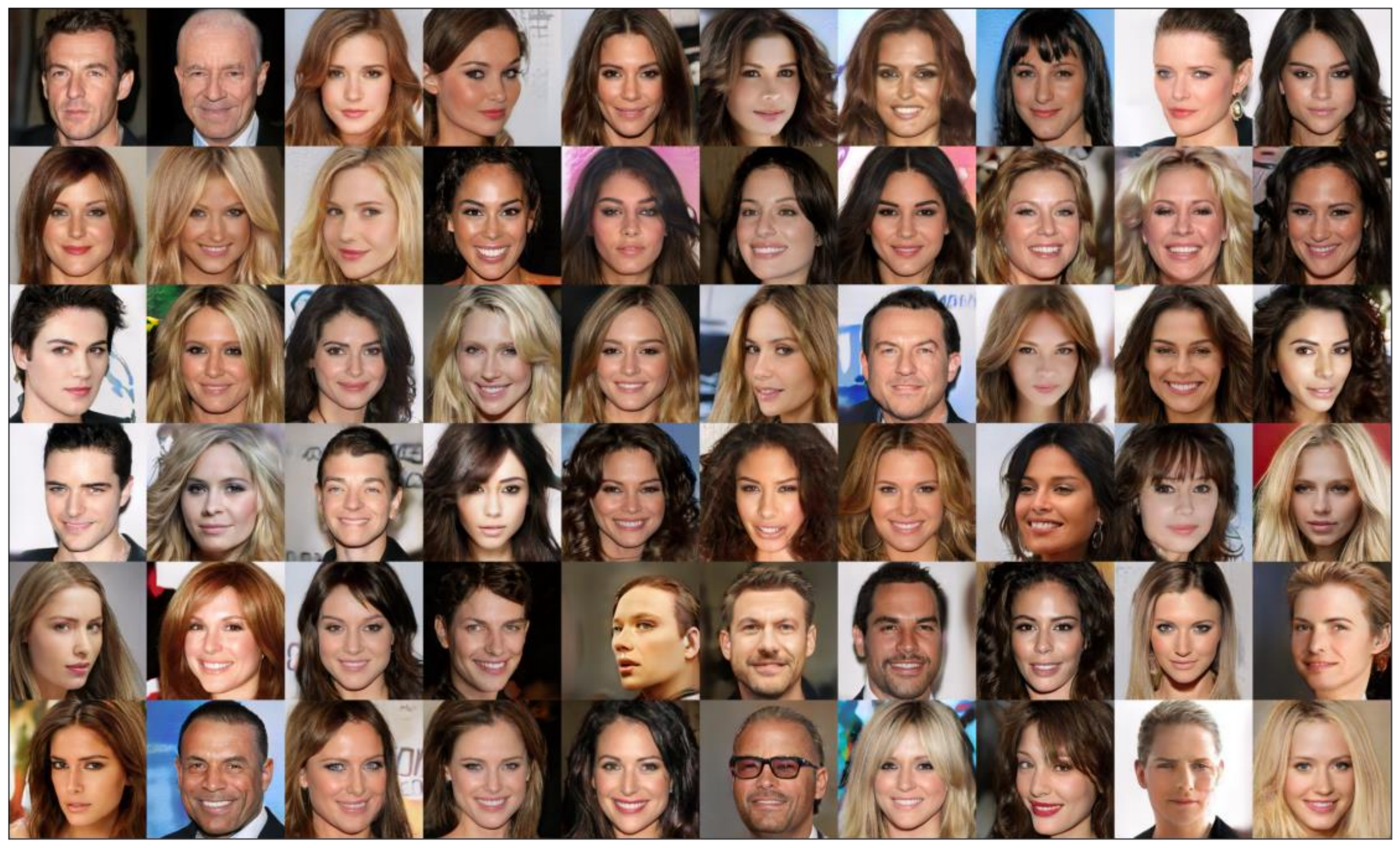}
         \caption{Bottom 25-84 samples in low CLA group.}
     \end{subfigure}
    \begin{subfigure}[b]{0.9\columnwidth}
         \centering
         \includegraphics[width=\textwidth]{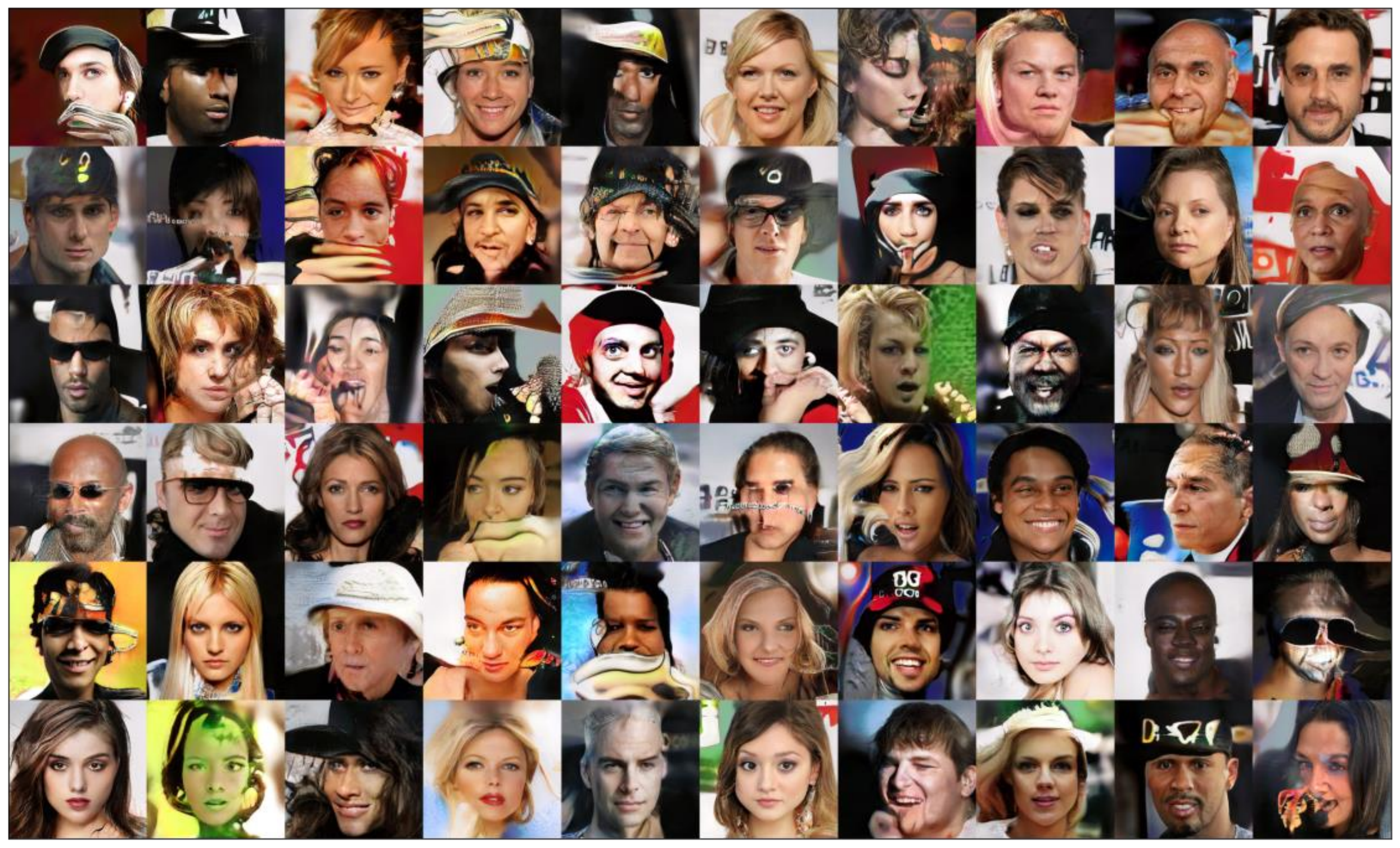}
         \caption{Top 25-84 samples in High CLA group.}
     \end{subfigure}

     \caption{Samples in the order of CLA in PGGAN with CelebA-HQ.}
     \label{fig:pggan_celeba}
\end{figure}

\section{Artifact Correction}
\begin{figure}[h!]
    \centering
         \centering
         \includegraphics[width=0.9\textwidth]{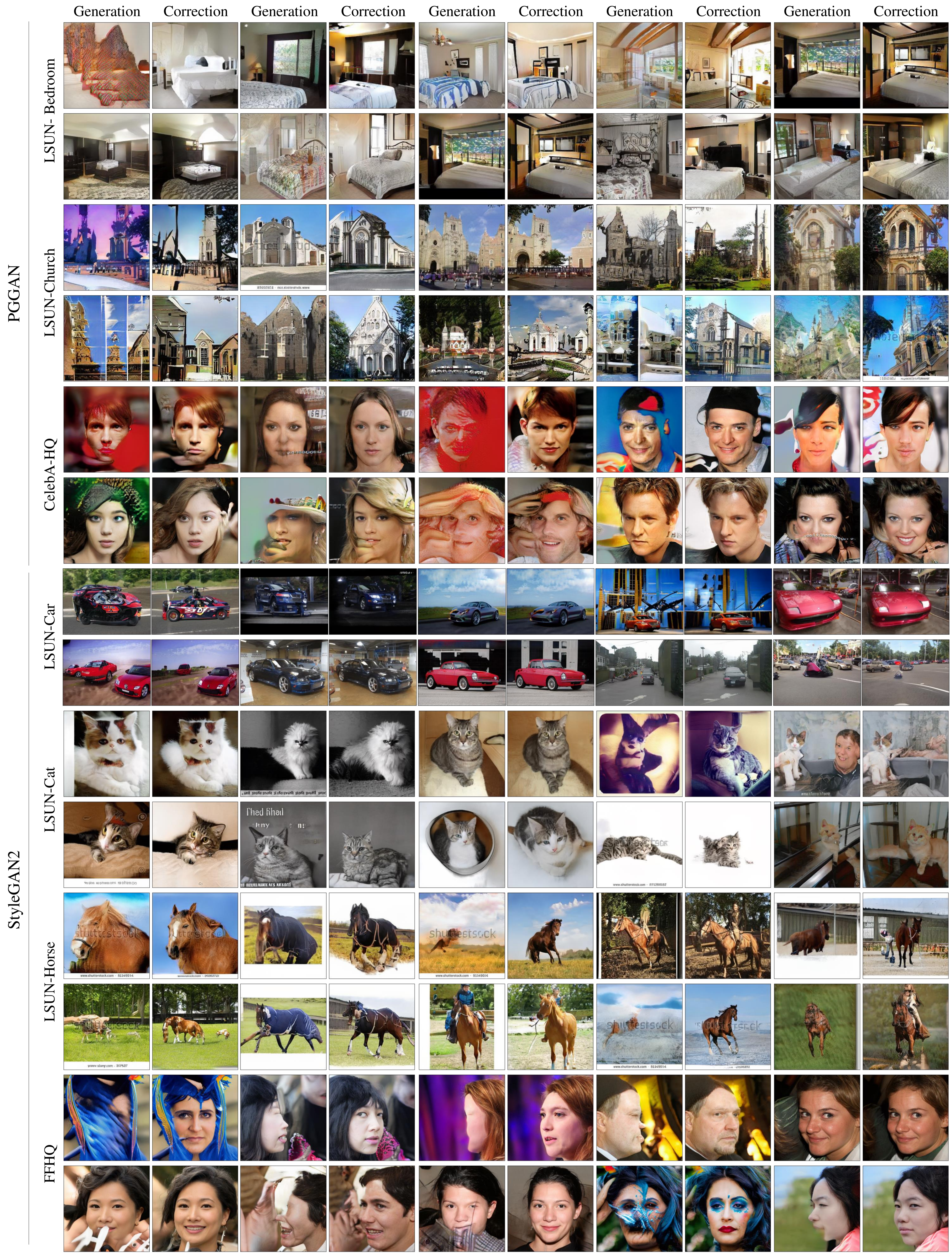}
         \caption{Examples for the correction results on various GANs for high CLA group.}
     \label{fig:corr_result}
\end{figure}
\end{document}